\documentclass[10pt,twocolumn,letterpaper]{article}

\usepackage{cvpr}              % To produce the CAMERA-READY version
\usepackage[accsupp]{axessibility}  % Improves PDF readability for those with disabilities.
\usepackage{enumitem}
\usepackage{amsmath,bm}
\usepackage{graphicx}

\usepackage[marginal]{footmisc}
\renewcommand{\thefootnote}{\fnsymbol{footnote}}
\usepackage[page]{appendix} % print appendices title

\usepackage{titletoc}

\definecolor{cvprblue}{rgb}{0.21,0.49,0.74}
\usepackage[pagebackref,breaklinks,colorlinks,allcolors=cvprblue]{hyperref}

\def\paperID{3134} % *** Enter the Paper ID here
\def\confName{ICCV}
\def\confYear{2025}

\title{Stable-Sim2Real: Exploring Simulation of Real-Captured 3D Data with Two-Stage Depth Diffusion}

\author{
Mutian Xu\textsuperscript{\rm 1} \quad Chongjie Ye\textsuperscript{\rm 2,3,1} \quad Haolin Liu\textsuperscript{\rm 4} \quad Yushuang Wu\textsuperscript{\rm 5} \quad Jiahao Chang\textsuperscript{\rm 1} \quad Xiaoguang Han\textsuperscript{\rm 1,2,3}\thanks{Corresponding author} \vspace{5pt}\\
\normalsize \textsuperscript{\rm 1}{SSE, CUHKSZ} \qquad \textsuperscript{\rm 2}{FNii-Shenzhen} \\
\normalsize \textsuperscript{\rm 3}{Guangdong Provincial Key Laboratory of Future Networks of Intelligence, CUHKSZ} \\
\normalsize \textsuperscript{\rm 4}{Tencent Hunyuan3D} \qquad
\textsuperscript{\rm 5}{ByteDance Games}
}
\begin{document}
\maketitle
\begin{abstract}
3D data simulation aims to bridge the gap between simulated and real-captured 3D data, which is a fundamental problem for real-world 3D visual tasks. Most 3D data simulation methods inject predefined physical priors but struggle to capture the full complexity of real data. An optimal approach involves learning an implicit mapping from synthetic to realistic data in a data-driven manner, but progress in this solution has met stagnation in recent studies.
% This work explores a new solution path of data-driven 3D simulation, via training a modern generative foundation model (\ie, SD: Stable Diffusion) on extensive synthetic-real paired data (\ie, LASA dataset).
This work explores a new solution path of data-driven 3D simulation, called Stable-Sim2Real, based on a novel two-stage depth diffusion model. The initial stage finetunes Stable-Diffusion to generate the residual between the real and synthetic paired depth, producing a stable but coarse depth, where some local regions may deviate from realistic patterns. To enhance this, both the synthetic and initial output depth are fed into a second-stage diffusion, where diffusion loss is adjusted to prioritize these distinct areas identified by a 3D discriminator. We provide a new benchmark scheme to evaluate 3D data simulation methods. Extensive experiments show that training the network with the 3D simulated data derived from our method significantly enhances performance in real-world 3D visual tasks. Moreover, the evaluation demonstrates the high similarity between our 3D simulated data and real-captured patterns. Project page: \href{https://mutianxu.github.io/stable-sim2real/}{mutianxu.github.io/stable-sim2real}.
\end{abstract}    
\section{Introduction}
\label{sec:intro}

% the 3D vision community has witnessed the bloom in training data-driven algorithms on \textit{real-world} 3D data to tackle diverse 3D vision and robotic tasks like 3D perception \cite{sunrgbd,s3dis,scannet,matterport3D}, reconstruction \cite{dtu,mvimgnet}, and manipulation \cite{anygrasp}. 

\begin{figure}[t]
  \centering
  \captionsetup{type=figure}
   \includegraphics[width=0.9\linewidth]{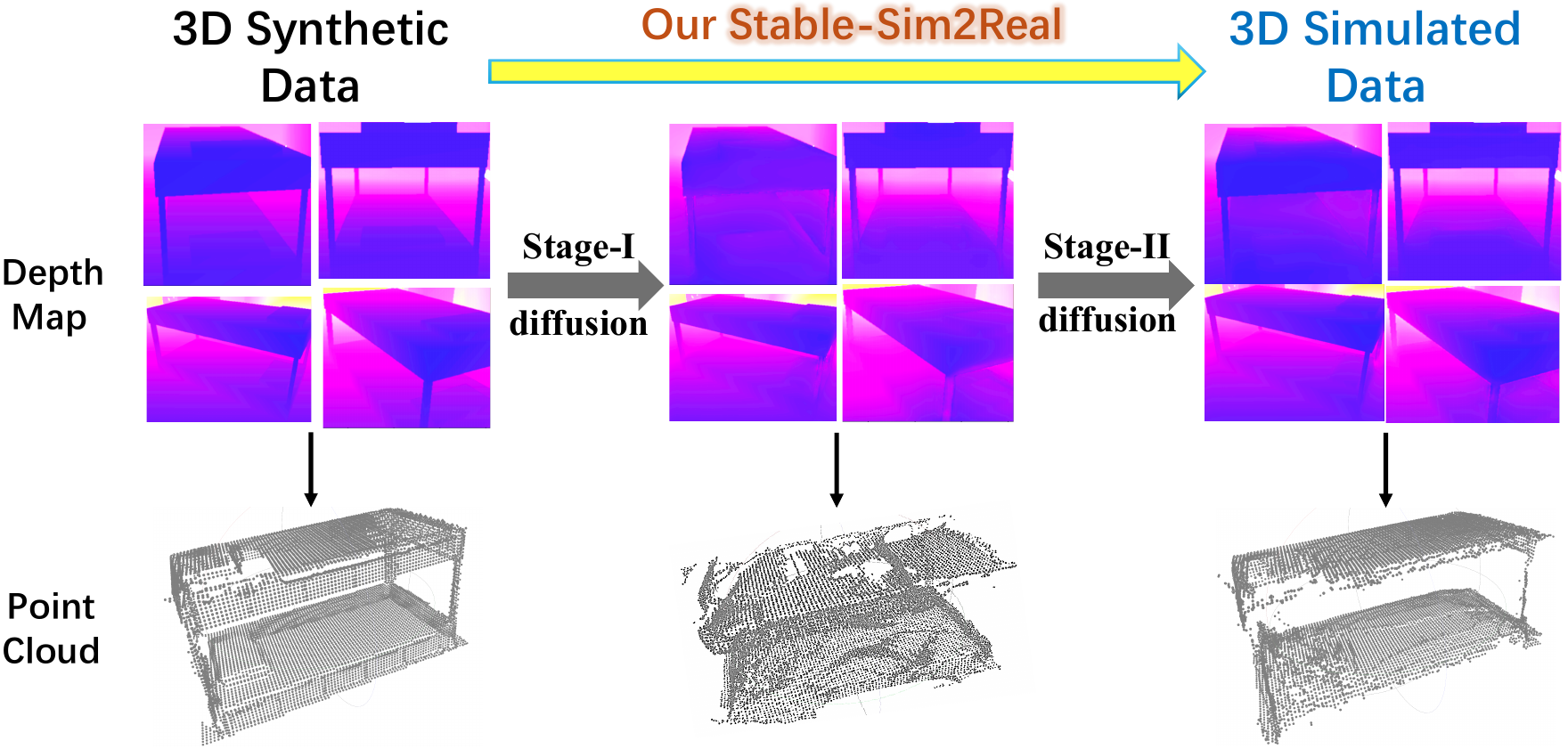}
   % \vspace{-0.2cm}
   \captionof{figure}{We delve back into a substantial yet stagnant problem of simulating real-captured 3D data in a data-driven manner, introducing \textit{Stable-Sim2Real}. 
   Given the depth maps of the 3D synthetic data, our approach applies a two-stage depth diffusion model based on Stable Diffusion \cite{stablediffusion} to generate realistic depth maps that are fused to produce real 3D point clouds. The initial stage generates stable but coarse depth maps, followed by a second-stage diffusion to enhance regional details.
   % Our model is trained on a large-scale synthetic-real paired dataset (\ie, LASA \cite{liu2024lasa}) in a data-driven manner.
   }
   \label{fig:teaser}
   \vspace{-0.3cm}
\end{figure}

In recent years, real-world 3D datasets have played a crucial role in addressing a wide range of tasks in 3D vision and robotics \cite{sunrgbd,s3dis,scannet,matterport3D,toscene,dtu,mvimgnet,anygrasp}.
% 3D perception \cite{sunrgbd,s3dis,scannet,matterport3D,toscene}, reconstruction \cite{dtu,mvimgnet}, and manipulation \cite{anygrasp}.
However, collecting real 3D data is often labor-intensive and time-consuming, and is further complicated by the recent concerns regarding data privacy.
% typically demands skilled users to perform scans using specialized equipment \cite{kinect,realsense}, a process that is often labor-intensive and time-consuming. Additionally, recent concerns regarding data privacy and usage rights further complicate real-world data acquisition. 
In this scenario, synthetic (\ie, simulated) data \cite{shapenet,modelnet,3dfront,abo,objaverse} emerges as an alternative data resource that is cost-effective, rapid, and scalable. Nevertheless, models trained using synthetic data may not perform robustly against the real-world signal.

This problem has catalyzed the development of 3D data simulation techniques, to minimize the gap between simulated and real-captured 3D data.
% Existing 3D data simulation methods primarily focus on enhancing the realism of \textit{depth} maps.
While some studies \cite{meister2013simulation,landau2015simulating,planche2017depthsynth,optical,light} have explored grounding physical priors to simulate depth sensors, they struggle to capture the full complexity of real-world patterns and are confined to simulation environments, due to the reliance on predefined and \textit{explicit} physical modeling.
% they are limited by the reliance on predefined and \textit{explicit} physical modeling. Consequently, they struggle to capture the full complexity of real-world data patterns and are confined to simulation environments.
A better solution involves learning an \textit{implicit} mapping from synthetic to realistic data in a data-driven manner, offering greater flexibility and adaptability to real-world diversities. 
However, only a handful of earlier works \cite{bousmalis2017unsupervised,sweeney2019supervised,gu2020coupled,shen2022dcl} have ventured down this path. Hindered by the lack of synthetic-real paired datasets and the dependence on conventional networks like GANs \cite{goodfellow2020generative}, which are prone to training instability and mode collapse \cite{gans}, these methods have yet to demonstrate the anticipated performance levels. As a result, developing data-driven 3D Sim2Real methods has long been a challenge within the community, and has \textit{shown little progress}.
% As a result, the adoption of data-driven 3D Sim2Real methods remains limited, leading to stagnation in related studies.
% despite the rapid demand for advanced 3D Sim2Real in robotics and automation \cite{sim2real_survey}, 

% Recently, a large-scale dataset, LASA \cite{liu2024lasa}, opens a new possibility to tackle this. It contains 10,412 high-quality 3D object CAD annotations aligned with the real-world instance scans from ARKitScenes \cite{baruch2021arkitscenes}. To utilize the power of LASA, and 
In this paper, we aim to explore the data-driven 3D Sim2Real and refocus the community's attention on this critical problem. As a prerequisite, we choose a recent synthetic-real paired dataset, LASA \cite{liu2024lasa}, containing 10,412 high-quality 3D shape CAD annotations aligned with the real-world object scans.
Driven by LASA, our efforts focus on designing an effective data-driven 3D Sim2Real algorithm. Given the inherent \textit{indeterminacy} and \textit{diversity} in real-captured data patterns, deep generative models emerge as a natural fit. Among these models, diffusion models \cite{song2019generative, song2020score, song2020improved,ho2020denoising,vahdat2021score} recently offer easier training dynamics and stronger generation ability.
% and have been extensively employed in 2D image \cite{diffbeat, meng2021sdedit, nichol2021glide, saharia2022image} and 3D shape generation \cite{luo2021diffusion, zhou20213d, 3dshapevec, qiu2024richdreamer}.
Yet, due to the lack of 3D data, it is hard to train a 3D diffusion model to get strong 3D priors for 3D simulation. Therefore, we opt to leverage the strong generalization priors of \textit{2D} diffusion foundation models (\eg, SD -- Stable Diffusion \cite{stablediffusion}) to simulate real 2D depth maps, which are then fused to gain 3D data. This strategy is similar to real 3D data acquisition, where 2D depth is captured and fused into 3D.
% Such an idea of using 2D diffusion model for 3D generation is commonly advocated in many 3D generation methods \cite{poole2023dreamfusion,zero123}.
% We advocate harnessing the strong generalization capabilities of the latent diffusion model, \ie, Stable Diffusion (SD) \cite{stablediffusion}. 

To this end, an intuitive baseline is to sample CAD (\ie, synthetic) depth maps and their paired real depth images from LASA, and then finetune SD to learn an implicit mapping between them. Although such a solution powered by large-scale paired data and the foundation diffusion model has shown effectiveness in conventional image translations, it faces \textbf{\textit{special challenges}} in our task: while conventional image translations (\eg, denoising/restoration) normally remove noise to produce \textit{clean} images, our task aims to output \textit{noisy} depth that is highly \textit{un}certain, thereby making the distribution that needs to be learned \textit{more complex}.
% \textit{fails} to simulate real noise

To address these challenges, we propose Stable-Sim2Real. The core is a novel two-stage depth diffusion model to simulate real-captured noise (a sample is shown in \cref{fig:teaser}). At Stage-I diffusion, instead of directly generating the corresponding real-world depth map, our model generates the \textit{residual} (\ie, difference) between real and CAD depth maps.
% focusing on a more concentrated distribution. 
The simulated depth map is then yielded by adding the generated residual with the CAD depth.
Empirically, this residual represents the noise added to clean (\ie, synthetic) depth during real-world scanning.
Moreover, compared to directly generating noisy real depth, adding noise to CAD depth, which is inherently clean and view-consistent, can produce more \textit{stable} depth with smaller view variation and better preservation of the original geometry (see \cref{sec:I} for more discussions and analysis).
% From experimental results, learning residual indeed performs better.
% However, in our experiment, a certain issue has emerged with this baseline model.
% However, according to our observation, the real-captured noise is spatially-\textit{variant}, \eg, different regions within a real depth map may exhibit varying degrees of noise. Nevertheless, our Stage-I diffusion learns towards adding spatially-\textit{uniform} noise across the entire input CAD depth, potentially due to the unfavorable bias within the diffusion model (similarly noted in \cite{mvedit2024}).

However, according to our observations and experiments, while some regions within the Stage-I generated depth map successfully fit the realistic pattern, some unsatisfactory areas still exhibit conspicuous 3D geometric discrepancies between the generation and real captures (shown in \cref{fig:lasa_sim_result_patch}).
To tackle this issue, our model then revolves around \textit{discriminating} unsatisfactory areas and \textit{enhancing} the Stage-I generation at local regions.
We train a 3D-Aware Discriminator to distinguish between Stage-I generation and real captures at local geometries, which is expected to delineate the \textit{distribution boundary} between them.
% We start by splitting all generated and real-captured depth maps into different patches to denote different areas, which are then projected to create 3D point patches. Then we introduce a point-based binary classifier to distinguish whether the input point patches come from model generation or real captures.
If a generated patch can be successfully discriminated, it signifies an unsatisfactory area.
% Next, we introduce a Stage-II diffusion, where local patches are sampled from Stage-II output and sent to the trained discriminator, constrained to be categorized as realistic patches.
Next, we introduce a Stage-II diffusion, conditioned not only on the CAD depth but also on Stage-I generation. 
More importantly, the diffusion loss is adjusted to \textit{prioritize} on unsatisfactory regions. In this way, Stage-II diffusion is \textit{\textbf{specialized}} on learning the distribution of unsatisfactory areas, thereby enhancing Stage-I generation.
Finally, the generated depth maps are fused to derive the simulated 3D data.
% \todo{In this way, Stage-II diffusion is guided to refine Stage-I generated depth to be \textit{locally closed} to real-world distributions.}
% After finding these patches, we further propose a second-stage depth refinement model, which differs from the baseline diffusion model in two key aspects. 
% , thereby reducing the inherent bias within SD
% \mt{Moreover, as the supervision signal is still derived from the data itself, no additional strong prior knowledge is introduced.} 
Notably, the binary classifier is only used during the training phase.

Last, we provide a comprehensive benchmark scheme for 3D data simulation.
The key logic is: if a model’s \textit{real-world} performance can be improved after being trained \textit{with simulated data}, it would validate the effectiveness of simulation methods. We focus on fundamental real-world 3D tasks: 3D shape reconstruction and 3D object/scene understanding.
% We apply our Stale-Sim2Real and other baseline methods on 3D synthetic datasets \cite{shapenet,future,abo,modelnet}, generating simulated 3D data.
As for shape reconstruction, we pretrain a 3D reconstruction network that takes simulated 3D data as input and outputs clean 3D surfaces. For 3D object/scene understanding, the simulated 3D data is used to pretrain a self-supervised point cloud learning framework. To better assess the performance gain purely from simulated data \textit{itself}, we conduct \textit{few}-shot evaluation on pre-trained networks.
 
Our contributions are summarized as:
\begin{itemize}[noitemsep,topsep=0pt]
	\item[$\bullet$] We delve back into a substantial yet stagnant problem, data-driven 3D real data simulation, paving a new solution path of training generative foundation models on extensive synthetic-real paired data.
        \item[$\bullet$] We present \textit{Stable-Sim2Real}, a novel 3D real data simulation method based on a two-stage depth diffusion model that can stably simulate real-captured noise.
        \item[$\bullet$] We provide a comprehensive scheme to benchmark 3D data simulation, based on few-shot evaluation on networks trained with simulated data for real-world 3D tasks.
	\item[$\bullet$] Extensive qualitative and quantitative results show that the data simulated by our method significantly boosts model performance on real-world 3D tasks. 
	% \item[$\bullet$] The code will be made publicly available.
\end{itemize}

\section{Related Work}
\label{sec:related_work}

\subsection{Simulation of Real-World Data}

\paragraph{Simulation of 2D real-world data.}
In the realm of 2D real-world data simulation, two predominant approaches have emerged: domain randomization and domain adaptation. The first line of works~\cite{random,tremblay2018training,tobin2018domain} randomly perturbs textures, lighting, and material of synthetic data, thereby modeling the synthetic-real domain gap as one kind of variation to endow models with robustness to real-world data. However, it brought the challenge of achieving a delicate balance between ensuring that the distribution of augmentations adequately covers that of real-world data and preventing the augmentation from being too extensive to hinder effective model training~\cite{kang2024balanced, yue2019domain, pashevich2019learning}. % Pashevich \etal~\cite{pashevich2019learning} propose to optimize the sequence of transformations for Sim2Real transfer.  
Domain adaptation, on the other hand, aims to minimize the domain gap by transferring knowledge from synthetic data to real-world scenarios, leveraging techniques such as adversarial training and self-supervised learning~\cite{shrivastava2017learning, bousmalis2018using, yuan2022sim}. 
% domain randomization and domain adaptation. \mt{(no need too many works, just list some representatives)}
% Domain randomization: The first line of works \cite{random,tremblay2018training} employs domain randomization to perturb textures, lighting and material of training data, so that the learned features can be robust to real-world data corruption. Yet, striking a balance between ensuring the augmentation distribution encompasses that of real data and not being too broad for model training for learning techniques poses a significant challenge.
% “Learning to Augment Synthetic Images for Sim2Real Policy Transfer”
% Domain adaptation: 
% "Learning from simulated and unsupervised images through adversarial training"
% "Using simulation and domain adaptation to improve efﬁciency of deep robotic grasping"
% "Sim-to-Real Transfer of Robotic Assembly with Visual Inputs Using CycleGAN and Force Control"
\paragraph{Simulation of 3D real-captured data.}
Acquiring real-world 3D data and their annotation is labor intensive and inefficient, deriving methods for the simulation of 3D data.
% such as RGB-D data~\cite{patricia2017deep,meister2013simulation,landau2015simulating,planche2017depthsynth,optical,light}, LiDAR data~\cite{denis2023gpu,tachella2019real,li2019aads,fang2020augmented,hossny2020fast}, and RADAR data~\cite{arnold2022maxray,schoffmann2021virtual,thieling2020scalable,schussler2021realistic,bialer2024radsimreal}, to serve 3D vision tasks including robotic grasping~\cite{torne2024reconciling}, autonomous driving~\cite{lin2022capturing, ding2022doda}, human pose estimation~\cite{martinez2018investigating}, and point cloud registration~\cite{chen2023sira}.
Existing simulation methods primarily focus on enhancing the realism of \textit{depth} information by grounding physical priors in simulators to close the domain gap with 3D data from the physical world~\cite{meister2013simulation,landau2015simulating,planche2017depthsynth,optical,light}. However, these approaches are usually limited by their reliance on predefined and explicit physical modeling. Alternatively, data-driven methods turn to implicitly learn from the distribution in real-captured data, independent of physical modeling~\cite{bousmalis2017unsupervised, sweeney2019supervised, gu2020coupled, shen2022dcl}, offering enhanced adaptability and generalization capabilities to various real-world scenarios. Nonetheless, due to the absence of extensive synthetic-real paired datasets and the reliance on traditional GANs \cite{goodfellow2020generative}, which suffer from training instability and mode collapse \cite{gans}, these approaches have not achieved the expected performance. Consequently, the research progress of data-driven 3D Sim2Real techniques faces a recent stagnation.

% Sensor-based: \cite{meister2013simulation,landau2015simulating,planche2017depthsynth,optical,light}

% Data-driven:
% “Deep depth domain adaptation: A case study,”
% “Investigating depth domain adaptation for efﬁcient human pose estimation"
% Chen \etal \cite{chen2023sira} inject pre-defined physical priors and data augmentation to simulate 3D data customized for 3D point cloud registration.
% \mt{more detailed and tell the pros and cons. Why data-driven?}
% \cite{bousmalis2017unsupervised, sweeney2019supervised, gu2020coupled, shen2022dcl}

\subsection{Image Generation and Translation.}
\paragraph{Conventional generative models.}
We formulate the 3D Sim2Real task as learning the mapping from synthetic to real-world captured noise, based on depth images. Existing approaches in the domain of image generation and translation include a variety of generative models such as Generative Adversarial Networks (GANs)~\cite{goodfellow2020generative}, Variational Autoencoders (VAEs)~\cite{kingma2014auto}, Cycle-Consistent Adversarial Networks (Cycle-GAN)~\cite{cyclegan}, and Pix2Pix~\cite{pix2pix}. While GAN-based methods, which have been predominant in image generation before diffusion models, achieved significant milestones in Sim2Real tasks~\cite{wang2020l2r, yuan2022sim, bresson2023sim2real}, they often struggle with maintaining local details and semantic consistency, leading to sub-optimal performance~\cite{gans, diffbeat, matsunaga2022fine, nichol2021improved, peng2023diffusion}.
% In contrast, our approach employs diffusion learning, a novel framework that addresses the weaknesses of traditional methods with a more stable training process and improved performance in generating high-quality images. By leveraging the strengths of diffusion learning, our method aims to enhance the realism of depth maps generated from simulated data, thereby reducing the distribution gap between simulated and real-world data.
% We formulate the 3D Sim2Real task as learning the mapping from synthetic depth map to real-world depth map.
% GAN, VAE, Cycle-GAN, Pix2Pix (selected as our main comparison), and finally diffusion. What are their weaknesses.
\paragraph{Diffusion models.} Diffusion models, also known as denoising diffusion probabilistic models (DDPMs)~\cite{dpm2015}, have emerged as a powerful alternative to GANs for image synthesis~\cite{diffbeat, meng2021sdedit, nichol2021glide, matsunaga2022fine, zhang2023adding, ruiz2023dreambooth, saharia2022image}. 
% These models operate by gradually refining a noisy input towards a clean image through a learned Markov chain process. 
Unlike GANs, diffusion models offer a more stable training procedure and have been shown to generate images with higher fidelity and better preservation of fine details~\cite{ho2022cascaded, stablediffusion}. Depth diffusion models recently track great attention. They mainly focus on the task of depth estimation in a monocular~\cite{saxena2023monocular, duan2025diffusiondepth, saxena2024surprising, tosi2024diffusion, ke2024repurposing, wang2024digging} or multi-view~\cite{wang2025mvdd, khan2021differentiable} manner. Other works also cover conventional image translation tasks, such as depth map super-resolution~\cite{shi2024dsr} and refinement~\cite{zhang2024betterdepth}, which produce \textit{clean} and \textit{certain} output images. Differently, our work leverages generative models to “generate \textit{noise}”, where the expected output is highly \textit{un}certain. Thus, the distribution that needs to be learned is more complex, bringing fundamentally different challenges.

% Why diffusion is more powerful? Diffusion for rgb image synthesis.

% Recent progress on depth diffusion models.
\section{Method}
\label{sec:method}

\begin{figure*}[ht]
    \centering
    \captionsetup{type=figure}
    \includegraphics[width=0.9\textwidth]{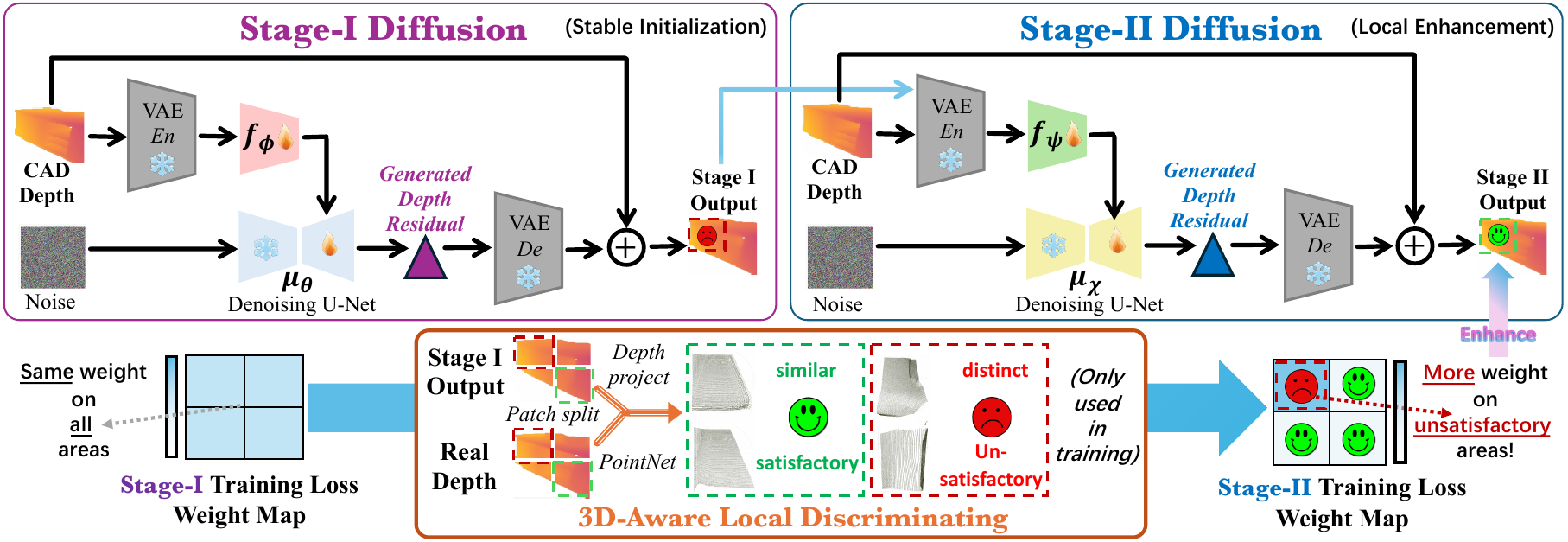}
    % \vspace{-0.6cm}
    \captionof{figure}{\textbf{An overview of our \textit{Stable-Sim2Real}}. Conditioned on CAD (\ie, synthetic) depth maps, our model first generates the \textit{residual} (\ie, difference) between real-captured and CAD depth maps. By adding the generated residual to the CAD depth, Stage-I diffusion produces stable but coarse depth maps (\cref{sec:I}). Next, the Stage-I output is projected into 3D and sent to a 3D discriminator, to identify the unsatisfactory local areas that are distinct from real-captured patterns (\cref{sec:discriminate}). Later, our Stage-II diffusion is conditioned on both CAD and Stage-I output, where the training loss of Stage-II is re-weighted to be specialized on enhancing unsatisfactory areas of Stage-I output (\cref{sec:stage2}). \textit{Notably}, the local discriminating and loss re-weighting are only used during training.
    % Finally, we fuse the output depth maps of Stage II to obtain the 3D simulated data.
    } \label{fig:pipeline}
    \vspace{-0.3cm}
\end{figure*}

We hope to exploit the strong prior of the diffusion-based foundation model, so we take \textit{depth} map as our bridge towards the simulation of real-captured 3D data. The preliminaries on the diffusion model are provided in \cref{sec:pre}.
Our Stable-Sim2Real is a two-stage depth diffusion framework, as illustrated in \cref{fig:pipeline}. The first-stage model (\cref{sec:I}) learns to generate the depth residual, producing stable but coarse depth maps, where some areas are distinct from the realistic pattern. Therefore, a 3D binary classifier (\cref{sec:discriminate}) is introduced to discriminate these distinct areas.
The second-stage model (\cref{sec:stage2}) is specialized to enhance the generation of these areas. Finally, the enhanced depth maps are fused to obtain the 3D simulated data (\cref{sec:fusion}).

\subsection{Preliminaries on Diffusion Models}\label{sec:pre}
Our model is based on Denoising Diffusion Probabilistic Models (DDPM) \cite{ho2020denoising,song2020denoising}.
The key of DDPM is to model a data distribution $p(x)$ via building a Markov chain in the data space. It can be divided into two main processes, forward process and backward process.

\paragraph{Forward diffusion process.} 
Given an initial data $x_{0}$, we gradually add random noise $\bm{\epsilon} \sim \mathcal{N}(\mathbf{0}, \mathbf{I})$ to $x_{0}$, yielding a series of noisy data, which can be simply formulated as $x_{t+1} = x_t + \bm{\epsilon}_t$.
After adding noise for multiple steps, the initial distribution of $p(x)$ is expected to be transformed into a standard Gaussian distribution.

According to DDPM \cite{ho2020denoising}, this process can be directly achieved by the following forward diffusion process:
\begin{equation}
q(x_t) = \sqrt{\alpha_t} {x_0} + \sqrt{1-\alpha_t} \boldsymbol{\epsilon}, x_0\sim p(x), t\in \{0, 1, ..., T\},
\label{eq:forward}
\end{equation} 
where $T$ denotes the number of the time step, t is the current time step, and $\alpha_t$ regulates the noise schedule deciding the pace at which the initial data distribution transforms into a standard Gaussian distribution.

\paragraph{Backward diffusion process.} 
Given a noisy $x_t$ from a standard Gaussian distribution as the initial state, ${x}_{t}$ is denoised to a cleaner $x_{t-1}$ at each time step, which can also be simply formulated as $x_{t-1} = x_t - \mu_\theta^{\bm{\epsilon}}(x_t, t) + \bm{\epsilon}_t$ in which $\bm{\epsilon}_t \sim \mathcal{N}(\mathbf{0}, \mathbf{\sigma_tI})$ and $\mu_\theta^{\bm{\epsilon}}$ is a learnable network to predict the injected noise. By repeating this reverse process until the maximum number of steps $T$, we can obtain the final generated data.

In DDPM \cite{ho2020denoising}, the backward diffusion process is proven to be:
\begin{align}
x_{t-1} = \frac{1}{\sqrt{\alpha_t}}x_t - \frac{1 - \alpha_t}{\sqrt{\alpha_t (1 - \Pi_{\tau=0}^t \alpha_\tau)}}\mu_\theta^{\bm{\epsilon}}(x_{t}, t;) + \sigma_t \bm{\epsilon},
\label{eq:backward}
\end{align}
where $\sigma_t$ denotes the standard deviation of $q(x_{t-1}|x_{t}, x_{0})$ in the backward diffusion to sample random noise. The loss for training $\mu_\theta^{\epsilon}$ is:
\begin{equation}
    L_{\theta} = \mathbb{E}_{\boldsymbol{\epsilon}, t} \left\| \boldsymbol{\epsilon} - {\mu_{\theta}}^{\boldsymbol{\epsilon}}(x_t, t)\right\|^2
\label{eq:denoise_loss_ori}
\end{equation}

\subsection{Stage-I: Initial Depth Generation}\label{sec:I}

\paragraph{Residual depth generation with SD.}
Different from unconditional diffusion models, text-to-image diffusion models focus on generating images with optional text prompts. A representative is Stable Diffusion (SD) \cite{stablediffusion}. SD trains a denoising U-Net architecture $\mu_\theta(x_t, t, c)$ on the latent space of a pre-trained Variational Auto-Encoder (VAE). Here, $c$ represents the supplementary text prompt embedding, usually acquired through methods like CLIP \cite{radford2021clip}.

We finetune SD model on synthetic-real paired depth data from LASA \cite{liu2024lasa}. Specifically, given an initial synthetic depth map $\boldsymbol{S}$ and its paired real-captured depth map $\boldsymbol{R}$ from a pre-defined dataset, we first get the initial ground truth residual (\ie, difference) depth map $\boldsymbol{D_{res}^{gt}} = \boldsymbol{S} - \boldsymbol{R}$.
Then we follow ControlNet \cite{controlnet} to add an additional encoder $En$ and decoder ${\boldsymbol{\mu}_{\theta}}^{\boldsymbol{\epsilon}}$ for finetuning SD.
Specifically, we run the forward process to add noise on $\boldsymbol{D_{res}^{gt}}$ following \cref{eq:forward} and encode with $En$ to obtain $x_t$.
Next, we send $\boldsymbol{S}$ to a pretrained VAE encoder $En$, generating the latent condition signal. This latent signal $En(\boldsymbol{S})$ is further processed by an additional encoder $\boldsymbol{f_\phi}$ and sent to the decoder blocks of denoising U-Net. We train both the decoder blocks ${\boldsymbol{\mu}_{\theta}}^{\boldsymbol{\epsilon}}$ and encoder $\boldsymbol{f_\phi}$ using the following loss (based on \cref{eq:denoise_loss_ori}):
\begin{equation}
L_{\theta, \phi} = \mathbb{E}_{\boldsymbol{\epsilon}, \boldsymbol{S}, t} \left\| \boldsymbol{\epsilon} - {\boldsymbol{\mu}_{\theta}}^{\boldsymbol{\epsilon}}(\boldsymbol{x}_t, t, \boldsymbol{f_{\phi}}(En(\boldsymbol{S}))) \right\|^2
\label{eq:denoise_stage1}
\end{equation}

\paragraph{Learning depth residual helps to \textit{produce stable depth.}}
Empirically, this residual represents the noise added to clean (\ie, synthetic) depth during real-world scanning.
Moreover, compared to directly generating noisy real depth, adding noise to CAD depth, which is inherently clean and view-consistent, can produce more \textit{stable} depth with smaller view variation and better preservation of the original geometry.
From experimental results, learning residual indeed performs better (see \cref{tab:quant_ablation}).

\paragraph{Learning depth residual is \textit{theoretically identical} to learning real depth.}
% In practical applications, depth measurements from sensors are inherently uncertain.
First, assume that real depth is a sample from a Gaussian distribution, denoted by $\boldsymbol{R} \sim \mathcal{N}(\mathbf{\alpha_{R}}, \mathbf{\vartheta})$, where $\alpha$ denotes the mean and $\vartheta$ represents the variance. Notably, $\vartheta$ can naturally represent real-captured noise.
Therefore, synthetic depth can be denoted by $\boldsymbol{S} \sim \mathcal{N}(\mathbf{\alpha_{S}}, 0)$, where the variance is zero as there is no real-captured noise.

In our design, as the diffusion model is inherently designed to learn the transformation from a random noise $\mathcal{N}(\mathbf{0}, \mathbf{I})$ to any other Gaussian distribution \cite{song2019generative}, we leverage it to model the distribution of the depth residual that can be denoted by $\mathcal{N}(\mathbf{\alpha_{R}-\alpha_{S}},\mathbf{\vartheta})$.
Last, by adding the depth residual to synthetic depth, the output distribution can be formulated as: $\mathcal{N}(\mathbf{\alpha_{R}-\alpha_{S}},\mathbf{\vartheta}) + \mathcal{N}(\mathbf{\alpha_{S}}, 0) = \mathcal{N}(\mathbf{\alpha_{R}}, \mathbf{\vartheta})$, which is exactly identical to the previously assumed real-captured noise distribution.

\subsection{3D-Aware Local Discriminating}
\label{sec:discriminate}

\paragraph{Unsatisfactory areas distinct from real patterns.}
Based on our observations, the noise captured in real data is spatially-\textit{variant}, meaning different areas within a real depth map may display varying levels of perturbation. However, our Stage-I diffusion tends to introduce spatially-\textit{uniform} noise across the entire input CAD depth. This discrepancy may stem from a bias within SD, as also discussed in \cite{mvedit2024}.
As a result, although certain regions of the generated depth map align well with real perturbation patterns, there remain noticeable areas that deviate from reality. This leads to evident 3D geometric discrepancies between Stage-I outputs and real captures (as illustrated in \cref{fig:lasa_sim_result_patch}).

\vspace{-0.1cm}
\paragraph{Identifying unsatisfactory areas with 3D discriminator.}
Our design logic to discriminate unsatisfactory areas is that if the generated areas diverge significantly from real patterns, there should exist a \textit{distribution boundary} of their \textit{3D} geometries. To this end, we propose employing a 3D binary classifier to discriminate this boundary.
Initially, we split all generated and real-captured depth maps into different patches representing different areas, which are then projected to form 3D point patches. Subsequently, we leverage PointNet \cite{qi2017pointnet} as a 3D binary classifier to differentiate whether the input point patches come from model generation or real captures. 
The CAD depth maps are also concatenated in the input as additional guidance.
The \textit{successful} classification of a generated patch indicates that its corresponding area is \textit{distinct} from real captures. \textit{Notably}, the binary classifier is solely used during training.

\subsection{Stage-II: Unsatisfactory Area Enhancement}\label{sec:stage2}

\paragraph{Improving unsatisfactory areas by re-weighting denoising loss.}
After identifying these local patches, we introduce a second-stage depth refinement model that differs from the Stage-I diffusion in two crucial aspects:
1) The diffusion model is now conditioned on the concatenation of synthetic depth $\boldsymbol{S}$ and the \textit{previously generated} depth residual $\boldsymbol{D}^{I}$ from Stage I. 
Here, $\boldsymbol{D}^{I}=\boldsymbol{D}_{res}^{I}+\boldsymbol{S}$, where $\boldsymbol{D}_{res}^{I}$ is the output residual depth from Stage-I model.
Thus, the training loss defined in \cref{eq:denoise_stage1} is updated to:
\begin{equation}
L_{\chi, \psi} = \mathbb{E}_{\boldsymbol{\epsilon}, \boldsymbol{D}^{I}, \boldsymbol{S}, t} \left\| \boldsymbol{\epsilon} - {\boldsymbol{\mu}_{\chi}}^{\boldsymbol{\epsilon}}(\boldsymbol{x}_t, t, \boldsymbol{f_{\psi}}(En(\boldsymbol{D}_{res}^{I}, \boldsymbol{S}))) \right\|^2,
\label{eq:denoise_stage2}
\end{equation}
2) More importantly, the diffusion loss is re-weighted to prioritize the enhancement of the identified unsatisfactory regions, making the final Stage-II loss $L_{\chi, \psi}$ to be:
\begin{equation}
L_{\chi, \psi} = \omega L_{\chi, \psi}^{sim} + \lambda L_{\chi, \psi}^{dis},
\label{eq:denoise_final}
\end{equation}
where $L_{\chi, \psi}^{sim}$ indicates the loss on similar regions, while $L_{\chi, \psi}^{dis}$ denotes the loss on previously identified distinct (\ie, unsatisfactory) areas. $\omega$ and $\lambda$ are weight coefficients, in which $\lambda > \omega$. In our experiment, setting $\omega=0.5$ and $\lambda=1.5$ produces the best result.

Our Stage-II diffusion model is \textbf{\textit{specialized}} on learning the distribution of unsatisfactory areas in Stage-I generation. Meanwhile, by conditioning on Stage-I output and assigning less weight to satisfied areas, the generation on satisfied outputs is also \textit{preserved} in Stage-II, as in \cref{fig:lasa_sim_result_patch}.

\subsection{Creating Real-World 3D Data} \label{sec:fusion}
\paragraph{Two-stage real depth simulation.} 
In the initial stage, we sample a random distribution $\bm{\epsilon}$ from $\mathcal{N}(\mathbf{0}, \mathbf{I})$ and feed it into the denoising U-Net $\boldsymbol{\mu}_{\theta}$ of Stage-I diffusion model. The model runs a backward diffusion process for $T$ steps to generate the initial depth residual $\boldsymbol{D}_{res}^{I}$, following \cref{eq:backward} with the synthetic depth $\boldsymbol{S}$ serving as the condition. 
Now we can obtain the initial depth map $\boldsymbol{D}^{I}=\boldsymbol{D}_{res}^{I}+\boldsymbol{S}$.

Moving to the second stage, we repeat a similar backward diffusion process to generate the residual depth $\boldsymbol{D}_{res}^{II}$, using the denoising U-Net $\boldsymbol{\mu}_{\chi}$ of our Stage-II diffusion model, conditioned on both synthetic depth $\boldsymbol{S}$ and Stage-I output $\boldsymbol{D}^{I}$. The final depth map $\boldsymbol{D}^{II}=\boldsymbol{D}_{res}^{II}+\boldsymbol{S}$.
DDIM sampler \cite{song2020denoising} is utilized during the whole inference phase.
% \textit{neither} a binary classifier \textit{nor} a re-weighted loss is used.

\paragraph{Depth fusion to 3D data.}
Given the 3D synthetic object, we follow LASA \cite{liu2024lasa} to embed CAD objects into 3D scenes. This allows us to render synthetic depth images, simulating the occlusions in real-world captures. Then we conduct the aforementioned two-stage real depth simulation on rendered synthetic depth maps, followed by a depth fusion algorithm to create simulated 3D data.

\section{Experiments}
\label{sec:experiments}

We provide a new benchmark scheme for 3D data simulation.
To begin with, different simulation methods are performed on synthetic datasets. The 3D simulated data are then utilized to train networks for tackling fundamental real-world 3D visual tasks: 3D instance reconstruction, 3D object/scene understanding. The \textbf{\textit{key logic}} of our evaluation is: if the model's real-world performance can be improved after being trained with simulated data, it would validate that the simulated data is \textit{close to real distributions}, demonstrating the effectiveness of simulation methods.

\begin{table*}[htbp]
\centering
\resizebox{0.9\linewidth}{!}{
\begin{tabular}{c|c|c|c|c|c|c}
\toprule
Method & Chair & Sofa & Table & Cabinet & Bed & Shelf \\
\midrule
w/o sim. data & 24.5 / 12.3 / 20.1 
& 55.6 / 7.02 / 19.8 
& 25.8 / 19.3 / 20.3 
& 57.1 / 8.83 / 24.1 
& 40.1 / 6.14 / 17.3 
& 12.8 / 6.74 / 23.7 \\ 
\midrule
Cycle-GAN \cite{cyclegan} & 25.1 / 12.1 / 20.9 
& 55.2 / 6.98 / 20.5 
& 26.1 / 19.1 / 19.5 
& 56.5 / 8.24 / 25.1 
& 40.6 / 5.98 / 16.9 
& 13.1 / 6.08 / 24.9\\
Pix2Pix \cite{pix2pix} & 24.9 / 11.0 / 20.5 
& 56.4 / 6.74 / 21.3 
& 26.6 / 18.5 / 22.1 
& 57.4 / 7.83 / 25.5 
& 41.3 / 6.01 / 17.8 
& 14.2 / 6.15 / 26.2 \\
Giga-GAN \cite{kang2023gigagan} & 26.2 / 10.3 / 22.1 
& 58.9 / 5.87 / 22.4 
& 29.3 / 16.4 / 24.3 
& 60.5 / 7.01 / 26.9 
& 44.2 / 5.25 / 19.3 
& 18.2 / 5.87 / 27.8 \\
Stable-Diffusion \cite{stablediffusion} & 27.3 / 9.51 / 25.5
& 59.4 / 5.93 / 23.6 
& 29.7 / 15.8 / 24.7 
& 61.6 / 6.92 / 27.5 
& 45.1 / 5.31 / 20.1 
& \textbf{19.1} / 5.85 / 28.3 \\
\textbf{Ours} & \textbf{29.2} / \textbf{7.83} / \textbf{27.5} 
& \textbf{62.9} / \textbf{4.09} / \textbf{25.1} 
& \textbf{31.2} / \textbf{14.1} / \textbf{26.8} 
& \textbf{63.2} / \textbf{5.58} / \textbf{28.6} 
& \textbf{46.3} / \textbf{4.23} / \textbf{21.8} 
& 18.8 / \textbf{5.02} / \textbf{31.6}\\
\bottomrule
\end{tabular}}
\caption{Quantitative comparison of few-shot evaluation on 3D instance reconstruction, where the model is pretrained using simulated data generated by different methods.  “w/o sim. data” indicates pretraining using \textit{no} simulated data but only synthetic data. Evaluation metrics are \textbf{mIoU / Chamfer L2 / F-score} respectively. Higher mIoU and F-score are better while lower Chamfer L2 is better. Chamfer L2 is scaled by 1,000. All results are reproduced using the official code of DisCo \cite{liu2024lasa}. \label{tab:quant_recon}}
\end{table*}

\begin{table*}[t]
\centering
\resizebox{0.86\linewidth}{!}{
\begin{tabular}{c|c|c|c|c|c|c}
\toprule
Method & Chair & Sofa & Table & Cabinet & Bed & Shelf \\
\midrule
w/o sim. data 
& 19.4 / 14.7 / 15.8 
& 46.2 / 9.92 / 15.7 
& 20.5 / 23.6 / 15.9 
& 51.2 / 11.3 / 20.2 
& 33.6 / 13.1 / 10.9 
& 11.1 / 9.23 / 21.6\\
\midrule
Cycle-GAN \cite{cyclegan} 
& 20.3 / 13.1 / 16.5 
& 48.9 / 9.03 / 18.9 
& 23.2 / 22.4 / 17.5 
& 51.9 / 10.8 / 21.4 
& 35.2 / 12.5 / 12.1 
& 11.7 / 8.87 / 23.5 \\
Pix2Pix \cite{pix2pix} 
& 19.3 / 12.3 / 17.4 
& 49.7 / 9.51 / 18.8 
& 22.6 / 21.5 / 18.8 
& 52.7 / 11.2 / 21.1 
& 37.7 / 10.0 / 13.8 
& 12.5 / 8.91 / 24.3 \\
Giga-GAN \cite{kang2023gigagan} 
& 22.4 / 11.8 / 19.5 
& 53.0 / 7.69 / 19.4 
& 23.1 / 19.3 / 21.6 
& 54.3 / 10.1 / 21.9 
& 40.3 / 7.95 / 15.4 
& 14.2 / 7.81 / 26.9 \\
Stable-Diffusion \cite{stablediffusion} 
& 23.3 / 11.7 / 20.4
& 54.1 / 7.98 / 19.2 
& 24.2 / 17.9 / 22.6
& 55.1 / 9.07 / 22.4 
& 40.4 / 7.89 / 16.2 
& 15.8 / 6.72 / 28.3\\
\textbf{Ours} 
& \textbf{27.1} / \textbf{9.31} / \textbf{24.8}
& \textbf{60.6} / \textbf{5.29} / \textbf{22.3} 
& \textbf{28.5} / \textbf{16.1} / \textbf{24.7} 
& \textbf{60.6} / \textbf{7.51} / \textbf{25.9} 
& \textbf{43.2} / \textbf{6.23} / \textbf{18.0} 
& \textbf{17.3} / \textbf{5.93} / \textbf{30.5}\\
\bottomrule
\end{tabular}}
\caption{Quantitative comparison of zero-shot evaluation on 3D instance reconstruction, when the model is solely pretrained using simulated data generated by different methods. 
% The notations are similar with \cref{tab:quant_recon}. 
\label{tab:quant_recon_simonly}
}
\vspace{-0.3cm}
\end{table*}

\subsection{Preliminaries}
\paragraph{Training of Stable-Sim2Real.}
We train our model by finetuning the official StableDiffusion  V2.1 \cite{stablediffusion}.
The training dataset is a recent 3D synthetic-real paired dataset, LASA \cite{liu2024lasa}. LASA is a large-scale dataset that contains 10,412 unique CAD models across 17 categories. Skilled artists are employed to meticulously craft aligned CAD models from 3D object real-world scans of \cite{baruch2021arkitscenes}. LASA provides CAD-real depth map pairs, rendered based on camera information during the real-world collection. See the supplementary material for more implementation details.

\paragraph{Methods in comparison.}
Since our method operates on depth images, we select several image-to-image generation methods for comparison:
1) Cycle-Consistent Adversarial Networks (Cycle-GAN) \cite{cyclegan}, which is a representative generative network when paired training data are not available, so we train Cycle-GAN on the \textit{pair-shuffled} LASA training data. 
2) Pix2Pix \cite{pix2pix}, which is a classical image generation method. 
3) Giga-GAN \cite{kang2023gigagan}, a state-of-the art GAN-based method.
4) Stable Diffusion \cite{stablediffusion}, our main baseline model.
% 5) Additionally, we add random noise to the synthetic depth map, indicating human-crafted perturbation for comparison.
Both 2), 3), 4) are trained using the paired data from LASA.
% For a clearer picture of our method, we also include the results of only using our Stage I model.

\subsection{3D Instance Reconstruction}\label{sec:experiments_recon}
\paragraph{Setup.}
3D instance reconstruction aims to recover the full geometry from real-captured partial point clouds or multi-view images. We utilize a recent state-of-the-art diffusion-based reconstruction network, DisCo, which is proposed in LASA \cite{liu2024lasa}. It is originally pretrained on three synthetic datasets (\ie, ShapeNet \cite{shapenet}, ABO \cite{abo}, and 3D-Future \cite{future}) and finetuned on LASA real training data.
During pretraining, DisCo simply constructs partial point clouds by adding random perturbation to synthetic data.
Instead, in our experiment, to simulate partial point clouds for pretraining, we perform simulation using our method and other baseline models based on the original three synthetic datasets used to pretrain DisCo, and replace half of the original partial point clouds with the simulated 3D data. The other configurations all follow the original DisCo.
% ~\footnote{\href{https://github.com/GAP-LAB-CUHK-SZ/DisCo}{github.com/GAP-LAB-CUHK-SZ/DisCo}}.

Subsequently, we employ few-shot and zero-shot learning configurations to evaluate: We randomly sample only a few LASA real training data to finetune the pretrained network. In this way, the real-world knowledge of the network will \textit{mostly} come from the pretraining data (\ie, the simulated data produced by our method and others) instead of the finetuning data, which can better reflect the performance gain from the simulated data \textbf{\textit{itself}}.
% The model is finetuned on 20\% samples and evaluated on the remaining validation samples.
The pretrained model is finetuned with 20$\sim$30\% of the original training samples for few-shot learning and is evaluated on the original LASA validation set.
For zero-shot learning, we directly evaluate the pretrained network without fine-tuning.
% 1) Mixing 3D simulated data with LASA: We mix the 3D simulated (partial) point clouds generated by our method and other approaches with the original LASA training data. The trained DisCo model is tested on LASA validation set, to evaluate the \textit{upper-bound} of model performance.
% 2) Only using 3D simulated data: We use the 3D simulated data independently, \textbf{without} mixing with the original LASA training data. The trained DisCo model is \textit{still} tested on LASA validation set, to directly evaluate the utility of the simulated data.

% \paragraph{Why not perform simulation on LASA synthetic data} and compare the performance using the simulated data with that using LASA paired real GT?
\textbf{Why not perform simulation on LASA synthetic data} and compare the performance using the simulated data with that using LASA paired real GT?
% We aim to leverage our method and other baseline techniques to generate 3D simulated data based on synthetic data. 
Since our StableSim2Real and other baseline techniques are trained on the LASA training set, it is unfeasible to conduct further inference on LASA, as this would not verify the actual model performance but only reflect the performance caused by overfitting. We can also not conduct simulation on a third-party paired dataset similar to LASA for 3D reconstruction, as there is currently \textit{no} such paired dataset.
% On the other hand, performing inference on the LASA validation set is also impractical due to the limited data quantity in the validation set (approximately 1000 samples), which may not provide adequate 3D simulated data for training reconstruction or classification networks.

\vspace{-0.3cm}
\paragraph{Quantitative results.}
Following \cite{liu2024lasa}, we utilize mean Intersection over Union (mIoU), L2 chamfer distance, and 1\% F-score metrics. Initially, we normalize both the results and the ground-truth CAD objects to fall within the range of -0.5 to 0.5, according to which we calculate the aforementioned metrics between them.
As shown in \cref{tab:quant_recon}, under all few-shot settings, the model pretrained using the simulated data from our method achieves the best performance, and significantly surpasses the model using no simulated data (\ie, only add random perturbation to synthetic data) for pretraining. 
% Additionally, our Stage I output can also attain similar or slightly better performance. 
Moreover, using the simulated data produced by Cycle-GAN or Pix2Pix does not improve the model performance compared to that using no simulated data, possibly attributed to their inferior simulation quality, introducing ambiguity and complexity during pretraining, making it hard for the model to learn useful real-world knowledge.
Moreover, as listed in \cref{tab:quant_recon_simonly}, under zero-shot setting, when solely using simulated data for pretraining without any finetuning, the model trained with our simulated data demonstrates more conspicuous effectiveness. 
% Additionally, under zero-shot setting, the performance gap between ours and other methods is conspicuously larger than it under few-shot setting, further demonstrating our effectiveness.

\begin{figure}[t]
  \centering
  \captionsetup{type=figure}
   \includegraphics[width=1.0\linewidth]{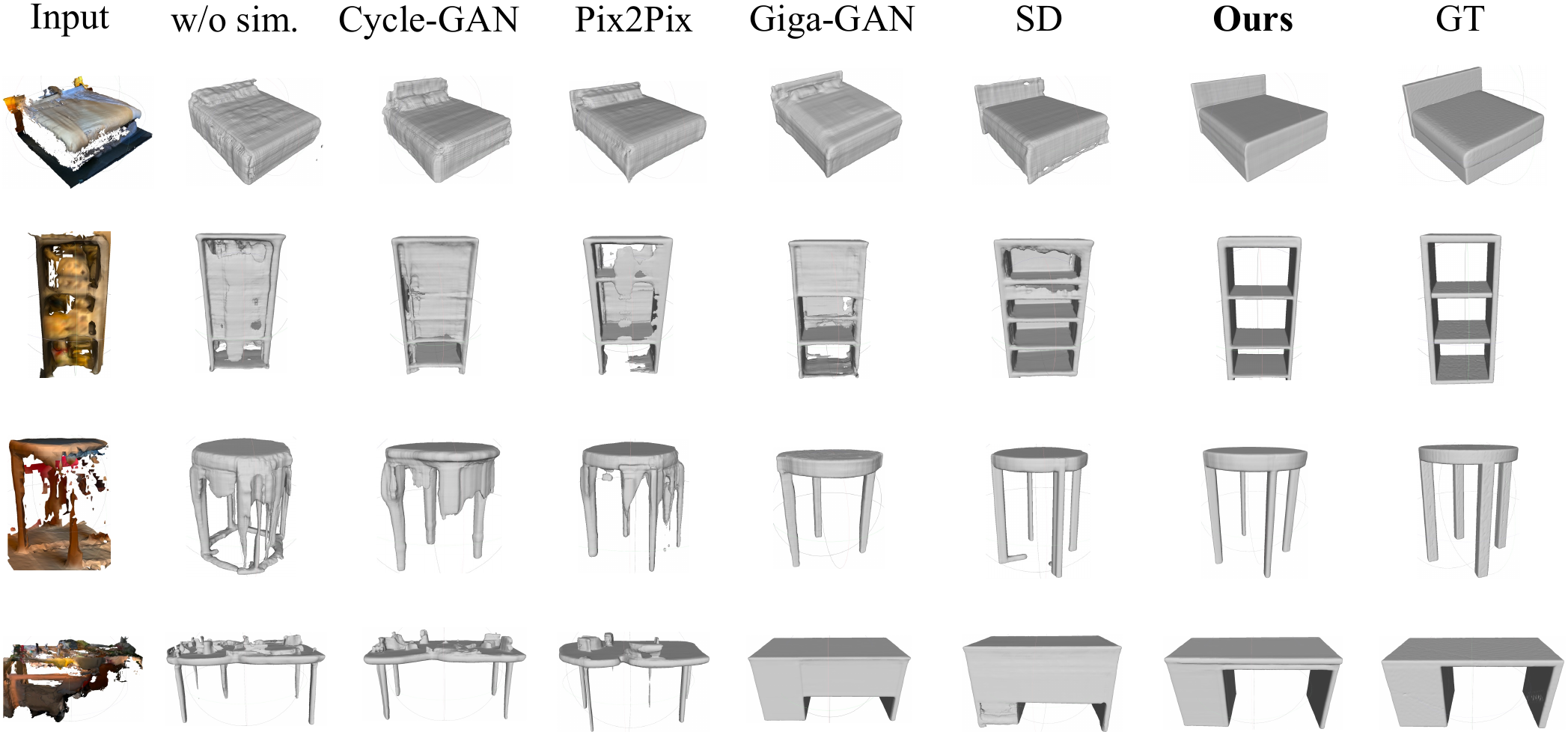}
   % \vspace{-0.2cm}
   \captionof{figure}{Qualitative comparison of few-shot evaluation on 3D instance reconstruction. “w/o sim.” denotes using \textit{no} simulated data for pretraining.}
   \label{fig:lasa_recon_result}
   \vspace{-0.3cm}
\end{figure}

\vspace{-0.3cm}
\paragraph{Qualitative results.}
\cref{fig:lasa_recon_result} shows the qualitative comparison, where our simulated data can help the model to gain the best reconstruction \textit{accuracy} and \textit{diversity}. 
% Moreover, given the reconstruction model DisCo is a diffusion-based model, we observe that our simulated data can enhance DisCo's proficiency in generating novel yet plausible geometries, such as the creation of a blanket on a bed.
More qualitative results are provided in the supplementary material.

\subsection{3D Object/Scene Understanding}

\paragraph{Setup.}
We select OcCo \cite{OcCo} as our backbone, which is a self-supervised point cloud pretraining framework. OcCo is originally pretrained on ModelNet \cite{modelnet}, and finetuned on real-scanned datasets -- ScanObjectNN \cite{scanobjectnn} for object classification, and S3DIS \cite{s3dis} for scene semantic segmentation. ModelNet contains 12,311 synthetic CAD objects from 40 classes. ScanObjectNN includes $\sim$15,000 real-scanned objects across 15 categories. S3DIS encompasses 271 indoor rooms and 13 classes. The classification task outputs the class of the whole 3D object, and the semantic segmentation predicts the category of each point in the 3D scene.

In our experiment, we apply our method and other baselines on ModelNet to generate simulated data that are further used to replace half of the original ModelNet synthetic data to pretrain OcCo.
Next, the pretrained model is again evaluated under the few-shot learning setting. A typical setting is “$K$-way $N$-shot”, which indicates that $K$ classes are randomly selected during finetuning, and each category contains $N$ samples.  The finetuned model is evaluated on the unseen samples picked from each of the $K$
classes. We select DGCNN \cite{dgcnn} as the backbone for OcCo. For few-shot learning, we employ the “5-way, 10-shot” setting. For S3DIS, we follow the original OcCo to report the results using 6-fold cross-validation. All the other configurations follow the original OcCo.

\vspace{-0.3cm}
\paragraph{Quatitative results.}
We report the classification accuracy and segmentation mean Intersection over Union (mIoU) in \cref{tab:quant_cls}. 
The model pretrained using our simulated data gets the best results across both 3D object classification and 3D scene understanding tasks. By pretraining with our simulated data, the model can acquire valuable real-world knowledge that helps downstream tasks. In contrast, pretraining with the low-quality simulated data from Cycle-GAN or Pix2Pix may even degrade the performance.

\begin{table}[t]
  \setlength{\tabcolsep}{3.0pt}
  \centering
  \resizebox{0.9\linewidth}{!}{
  \begin{tabular}{c||c|c|c|c|c|c}
    \toprule
    Method & w/o sim. & C-GAN~\cite{cyclegan} & P2P~\cite{pix2pix} &  G-GAN~\cite{kang2023gigagan} & SD \cite{stablediffusion} & \textbf{Ours} \\
    \midrule
    Acc. (\%) & 72.4 & 72.5 & 72.1 & 74.2 & 74.6 & \textbf{77.5}\\
    \midrule
    mIoU (\%) & 47.9 & 48.0 & 47.7 & 49.1 & 48.9 & \textbf{51.2}\\
     % & 72.4 & 48.9\\
    % \midrule
    % Cycle-GAN~\cite{cyclegan} & 72.1 & 49.0\\
    % Pix2Pix~\cite{pix2pix} & 72.5 & 48.9\\
    % Giga-GAN~\cite{kang2023gigagan} & 74.2 & 50.1\\
    % Stable-Diffusion~\cite{stablediffusion} & 74.6 & 49.8 \\
    % \textbf{Ours} & \textbf{78.1} & \textbf{51.9} \\
    \bottomrule
  \end{tabular}}
  \vspace{-0.2cm}
  \caption{Quantitative comparison of few-shot evaluation on 3D object classification and 3D scene semantic segmentation. “w/o sim” means using \textit{no} simulated data for pretraining.} 
  \label{tab:quant_cls}
  \vspace{-0.1cm}
\end{table}

\subsection{Is the simulated data close to real captures?}

\begin{figure}[t]
  \centering
  \captionsetup{type=figure}
   \includegraphics[width=1.0\linewidth]{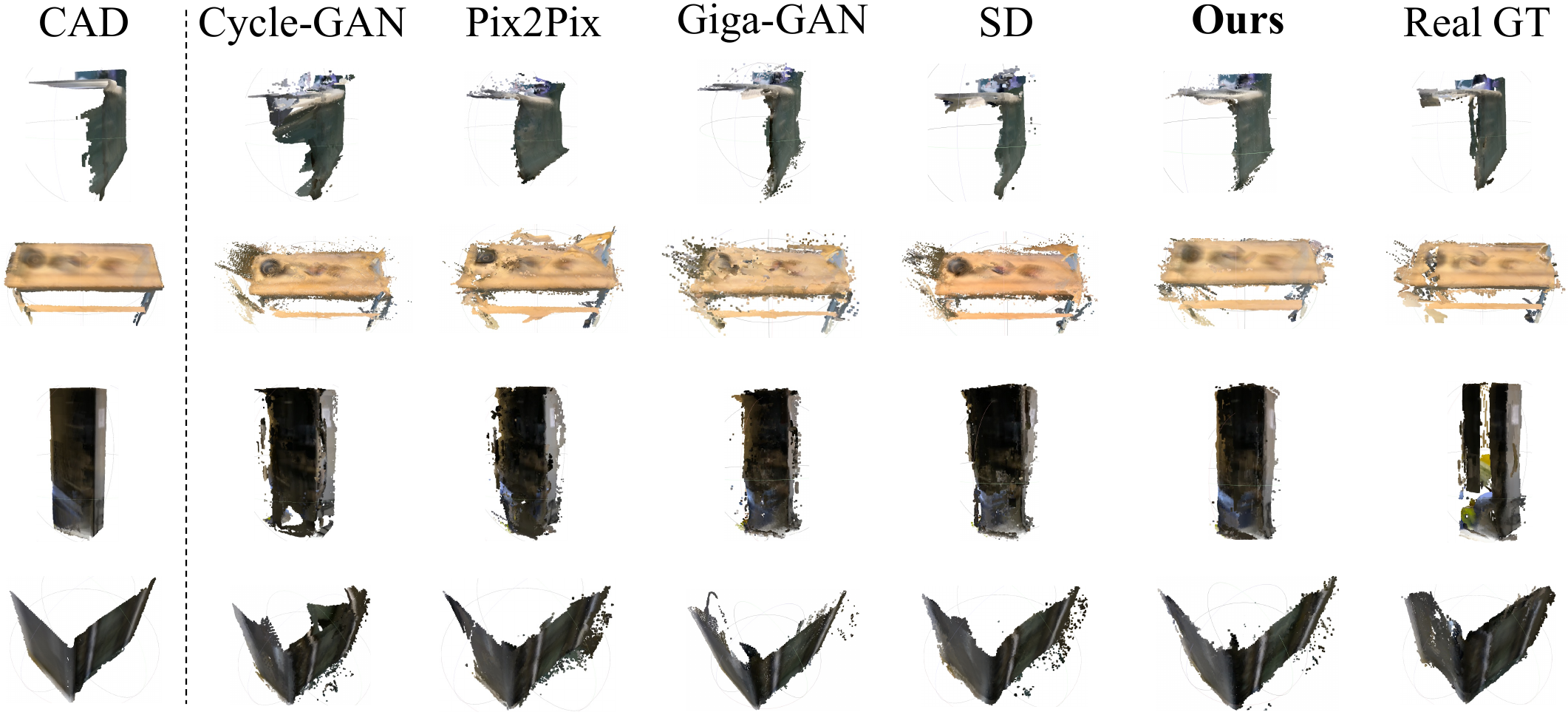}
   % \vspace{-0.2cm}
   \captionof{figure}{Qualitative simulation results on LASA \cite{liu2024lasa}.}
   \label{fig:lasa_sim_result}
   \vspace{-0.1cm}
\end{figure}

% \paragraph{Evaluation on LASA validation set using ground truth real-world captures.}
As previously mentioned, our Stable-Sim2Real is trained on LASA training set. In this section, we aim to evaluate the data simulation quality by directly performing inference and evaluation on LASA validation set.
\cref{tab:quant_cd} compares the mIoU, Chamfer L2, and F-score metrics between the simulated data generated by different methods and the real captured Ground Truth (GT). Although not precise, this result can partially prove that our simulated data closely resembles real-world captures.
\cref{fig:lasa_sim_result} shows simulation results on depth maps and 3D point clouds. It demonstrates that while our simulated data closely aligns with real-world captures, \textit{subtle differences} persist, highlighting the diversity.

\begin{table}[t]
  \setlength{\tabcolsep}{3.0pt}
  \centering
  \resizebox{1.0\linewidth}{!}{
  \begin{tabular}{c||c|c|c|c|c|c}
    \toprule
    Method & w/o sim. & C-GAN~\cite{cyclegan} & P2P~\cite{pix2pix} &  G-GAN~\cite{kang2023gigagan} & SD \cite{stablediffusion} & \textbf{Ours} \\
    \midrule
     Metrics & 59.2/3.54/25.3 & 60.4/3.21/25.2 & 61.5/3.02/26.4 
     & 62.1/2.86/26.7 & 62.3/2.83/27.5 & \textbf{62.9/2.57/28.2}\\
    \bottomrule
  \end{tabular}}
  \vspace{-0.2cm}
  \caption{The \textbf{mIoU / Chamfer L2 / F-score} between simulated data and real-captures on LASA \cite{liu2024lasa} validation set.} 
  \label{tab:quant_cd}
  \vspace{-0.3cm}
\end{table}

\begin{figure}[t]
  \centering
  \captionsetup{type=figure}
   \includegraphics[width=1.0\linewidth]{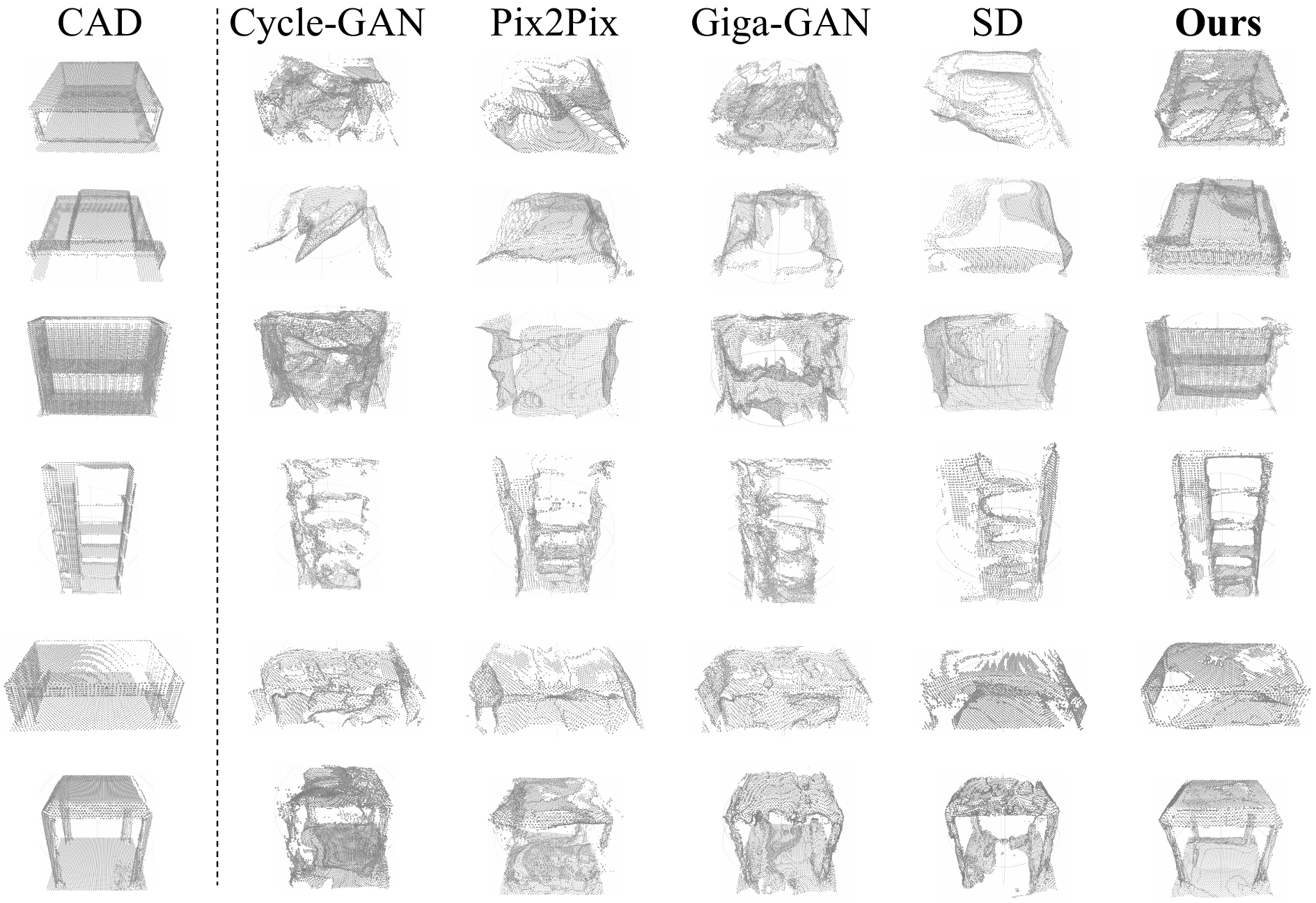}
   % \vspace{-0.2cm}
   \captionof{figure}{Qualitative simulation results on ShapeNet \cite{shapenet}.}
   \label{fig:syn_sim_result}
   \vspace{-0.3cm}
\end{figure}

We also provide the simulation results on ShapeNet \cite{shapenet} in \cref{fig:syn_sim_result}. Despite the absence of real-captured Ground Truth (GT) in ShapeNet, the outcomes still show that the simulated results from our output closely resemble real captures, where the generated geometry is not largely broken, with certain realistic perturbations introduced. 

\subsection{Ablation Studies and Analysis}

\paragraph{Module impacts.}
To evaluate the impact of each key design in our framework, we conduct ablation studies on the 3D instance reconstruction task, specifically on the table category. Similar to \cref{tab:quant_recon_simonly}, we conduct experiments under zero-shot protocol. This allows for a focused evaluation of the module's impact without the influence of additional factors. 
% using only the 3D simulated data independently, without mixing it with the original LASA training data.
Other setups remain consistent with \cref{sec:experiments_recon}.

% The quantitative results of using different modules are reported in .
1) \textit{Residual generation}. In \cref{tab:quant_ablation} (w/o res.), directly generating the real-world depth map without using residual generation degrades the model performance, verifying the efficacy of our residual generation.

2) \textit{Loss re-weighting}. In our Stage-II model, we re-weight the diffusion loss to focus more on unsatisfactory areas. In \cref{tab:quant_ablation} (w/o re-w.), the result w/o loss re-weighting is provided, showing its effectiveness.
% of our loss re-weighting operation.

3) \textit{3D binary discriminator}. Instead of training a 3D binary discriminator, another simple way is to directly set a threshold of the absolute difference between Stage-I generation and real-world GT to decide distinct areas. We tried various difference thresholds and reported the best result in \cref{tab:quant_ablation} (w/o bin.), which hurts the model performance.
To explain, our output depth is gained by adding \textit{subtle} noise to synthetic depth. Thus, the \textit{magnitude} of the difference between our output and its corresponding real GT is \textit{similar} at 2D/pixel space, where it is impractical to find an appropriate threshold to identify distinct/unsatisfactory areas.
Instead, their difference in distribution space is relatively more significant, so using a 3D binary classifier to delineate their \textit{\textbf{distribution} boundary} is a more feasible way.

4) \textit{Number of stages}. Theoretically, we can iteratively improve the generation quality by adding more stages. However, as shown in \cref{tab:quant_ablation}, adding Stage-III model does not improve the model performance, and Stage-IV model even slightly degrades the result. To analyze, Stage-II model has been sufficient to identify most of the unsatisfactory areas and improve their generation quality. In addition, adding redundant stages may also cause overfitting issues that hurt the performance.

\begin{table}[t]
  \setlength{\tabcolsep}{3.0pt}
  \centering
  \resizebox{1.0\linewidth}{!}{
  \begin{tabular}{c|c|c|c|cc|c}
    \toprule
    Method & w/o res. & w/o re-w. & w/o bin. & Stage-III & Stage-IV & \textbf{Stage-II}\\
    \midrule
     Metrics & 26.4/18.2/22.1 & 27.1/17.9/23.3 & 27.3/17.5/23.6 & \textbf{28.6}/
    \textbf{16.1}/24.5 & 28.0/16.9/24.1 & 28.5/\textbf{16.1}/\textbf{24.7}\\
    \bottomrule
  \end{tabular}}
  % \vspace{-0.2cm}
  \caption{Quantitative ablations of zero-shot evaluation on 3D instance reconstruction of table category. Evaluation metrics are \textbf{mIoU / Chamfer L2 / F-score} respectively.}
  \label{tab:quant_ablation}
  % \vspace{-0.3cm}
\end{table}

\begin{figure}[t]
  \centering
  \captionsetup{type=figure}
   \includegraphics[width=0.7\linewidth]{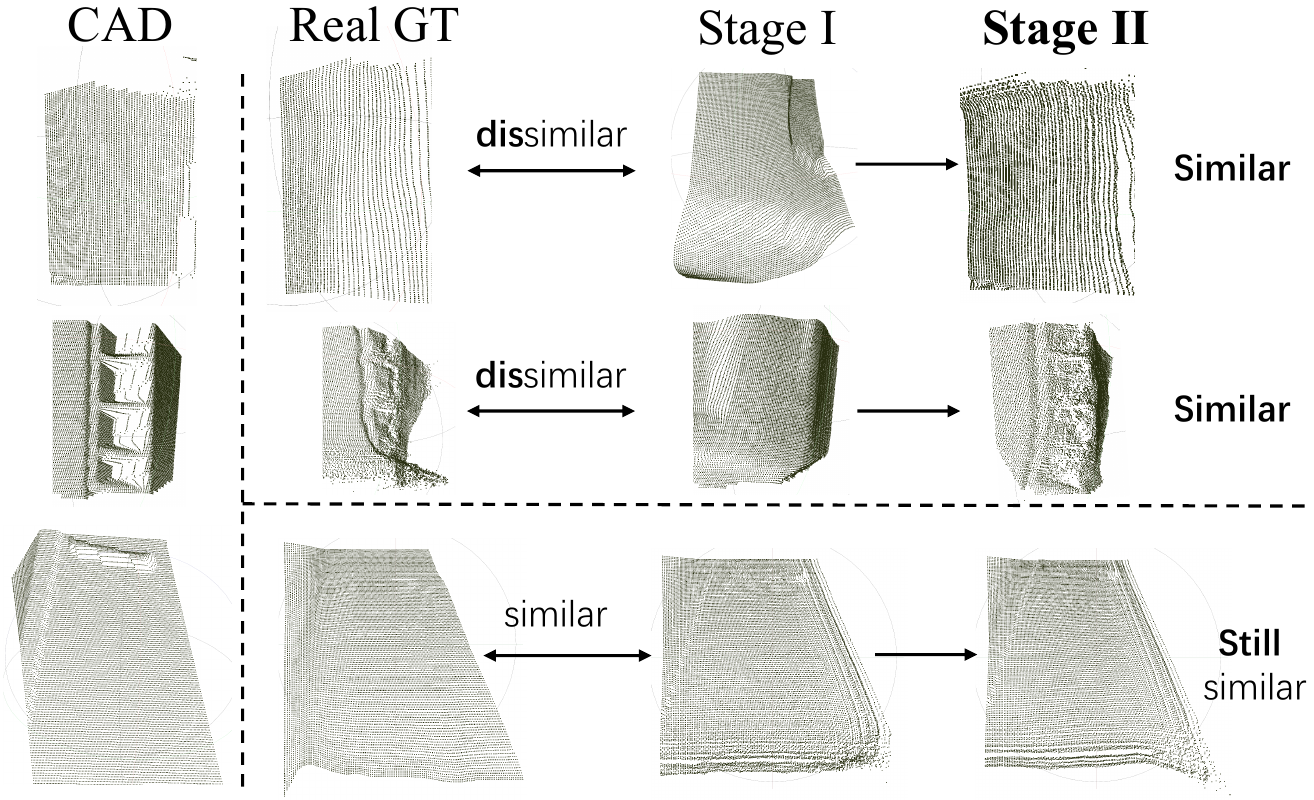}
   % \vspace{-0.2cm}
   \captionof{figure}{Qualitative simulation results at the patch level.}
   \label{fig:lasa_sim_result_patch}
   \vspace{-0.3cm}
\end{figure}

\vspace{-0.3cm}
\paragraph{Analyzing Stage-II model at \textit{3D patch}-level.}
1) We compare simulation results with real-captured Ground Truth (GT) at the patch level.
% to reflect the transition from Stage-I output to Stage-II output more intuitively.
As illustrated in \cref{fig:lasa_sim_result_patch}, our Stage-II model effectively refines previously unsatisfactory (\ie, dissimilar to GT) patches from Stage-I output to be similar to GT, while keeping previously satisfactory (\ie, similar to GT) patches almost unchanged.

2) % In our approach, the 3D binary classifier serves as the \textit{distribution} discriminator between the model-generated outputs and real captures. 
When sending the newly generated output patches from Stage-II to our 3D binary discriminator, the classification accuracy notably \textit{drops} from $\mathbf{67.8\%}$ to $\mathbf{32.5\%}$, suggesting that the generated patches successfully \textit{approximate} the real-captured distributions, making it challenging for the classifier to discriminate. This again verifies the effectiveness of our Stage-II model.
See the supplementary material for more ablations of the binary classifier.

% Rather than performing evaluation on single-depth-level tasks such as depth enhancement in previous data-driven 3D Sim2Real methods \cite{sweeney2019supervised,gu2020coupled,shen2022dcl} are 
\section{Conclusion}
\label{sec:conclusion}

We introduce Stable-Sim2Real, a two-stage depth diffusion model for 3D real data simulation. Initially, our method learns the residual between synthetic and real depth to generate a stable but coarse depth, where some areas may deviate from realistic patterns. Next, both synthetic and initial output depth are fed into the second-stage model, where diffusion loss prioritizes enhancing distinct regions identified by a binary classifier. Experiments show that the 3D simulated data generated from our method can improve the network performance in real-world 3D tasks, and our simulated data is highly similar to real captures. More limitations are discussed in the supplementary material.

\section*{Acknowledgement}
The work was supported in part by Guangdong S\&T Programme with Grant No.~2024B0101030002, the Basic Research Project No.~HZQB-KCZYZ-2021067 of Hetao Shenzhen-HK S\&T Cooperation Zone, the Shenzhen Outstanding Talents Training Fund 202002, the Guangdong Research Projects No.~2017ZT07X152 and No.~2019CX01X104, the Guangdong Provincial Key Laboratory of Future Networks of Intelligence (Grant No.~2022B1212010001), and the Shenzhen Key Laboratory of Big Data and Artificial Intelligence (Grant No.~ZDSYS201707251409055).

{
    \small

    \bibliographystyle{ieeenat_fullname}
    % \bibliography{main}

\begin{thebibliography}{80}
\providecommand{\natexlab}[1]{#1}
\providecommand{\url}[1]{\texttt{#1}}
\expandafter\ifx\csname urlstyle\endcsname\relax
  \providecommand{\doi}[1]{doi: #1}\else
  \providecommand{\doi}{doi: \begingroup \urlstyle{rm}\Url}\fi

\bibitem[{Armeni} et~al.(2016){Armeni}, {Sener}, {Zamir}, {Jiang}, {Brilakis}, {Fischer}, and {Savarese}]{s3dis}
I. {Armeni}, O. {Sener}, A.~R. {Zamir}, H. {Jiang}, I. {Brilakis}, M. {Fischer}, and S. {Savarese}.
\newblock 3d semantic parsing of large-scale indoor spaces.
\newblock In \emph{CVPR}, 2016.

\bibitem[Bai et~al.(2024)Bai, Zhang, Chen, Wan, and Zhang]{light}
Kaixin Bai, Lei Zhang, Zhaopeng Chen, Fang Wan, and Jianwei Zhang.
\newblock Close the sim2real gap via physically-based structured light synthetic data simulation.
\newblock In \emph{ICRA}, 2024.

\bibitem[Baruch et~al.(2021)Baruch, Chen, Dehghan, Feigin, Fu, Gebauer, Kurz, Dimry, Joffe, Schwartz, and Shulman]{baruch2021arkitscenes}
Gilad Baruch, Zhuoyuan Chen, Afshin Dehghan, Yuri Feigin, Peter Fu, Thomas Gebauer, Daniel Kurz, Tal Dimry, Brandon Joffe, Arik Schwartz, and Elad Shulman.
\newblock {ARK}itscenes: A diverse real-world dataset for 3d indoor scene understanding using mobile {RGB}-d data.
\newblock In \emph{NeurIPS}, 2021.

\bibitem[Bousmalis et~al.(2017)Bousmalis, Silberman, Dohan, Erhan, and Krishnan]{bousmalis2017unsupervised}
Konstantinos Bousmalis, Nathan Silberman, David Dohan, Dumitru Erhan, and Dilip Krishnan.
\newblock Unsupervised pixel-level domain adaptation with generative adversarial networks.
\newblock In \emph{CVPR}, 2017.

\bibitem[Bousmalis et~al.(2018)Bousmalis, Irpan, Wohlhart, Bai, Kelcey, Kalakrishnan, Downs, Ibarz, Pastor, Konolige, et~al.]{bousmalis2018using}
Konstantinos Bousmalis, Alex Irpan, Paul Wohlhart, Yunfei Bai, Matthew Kelcey, Mrinal Kalakrishnan, Laura Downs, Julian Ibarz, Peter Pastor, Kurt Konolige, et~al.
\newblock Using simulation and domain adaptation to improve efficiency of deep robotic grasping.
\newblock In \emph{ICRA}, 2018.

\bibitem[Bresson et~al.(2023)Bresson, Xing, and Guo]{bresson2023sim2real}
Marc Bresson, Yang Xing, and Weisi Guo.
\newblock Sim2real: Generative ai to enhance photorealism through domain transfer with gan and seven-chanel-360°-paired-images dataset.
\newblock \emph{Sensors}, 24\penalty0 (1):\penalty0 94, 2023.

\bibitem[Chang et~al.(2017)Chang, Dai, Funkhouser, Halber, Niessner, Savva, Song, Zeng, and Zhang]{matterport3D}
Angel Chang, Angela Dai, Thomas Funkhouser, Maciej Halber, Matthias Niessner, Manolis Savva, Shuran Song, Andy Zeng, and Yinda Zhang.
\newblock Matterport3d: Learning from rgb-d data in indoor environments.
\newblock In \emph{3DV}, 2017.

\bibitem[Chang et~al.(2015)Chang, Funkhouser, Guibas, Hanrahan, Huang, Li, Savarese, Savva, Song, Su, et~al.]{shapenet}
Angel~X Chang, Thomas Funkhouser, Leonidas Guibas, Pat Hanrahan, Qixing Huang, Zimo Li, Silvio Savarese, Manolis Savva, Shuran Song, Hao Su, et~al.
\newblock Shapenet: An information-rich 3d model repository.
\newblock \emph{arXiv preprint arXiv:1512.03012}, 2015.

\bibitem[Chen et~al.(2024)Chen, Shi, Liu, Shen, Gu, Wetzstein, Su, and Guibas]{mvedit2024}
Hansheng Chen, Ruoxi Shi, Yulin Liu, Bokui Shen, Jiayuan Gu, Gordon Wetzstein, Hao Su, and Leonidas Guibas.
\newblock Generic 3d diffusion adapter using controlled multi-view editing.
\newblock \emph{arXiv preprint arXiv:2403.12032}, 2024.

\bibitem[Collins et~al.(2022)Collins, Goel, Deng, Luthra, Xu, Gundogdu, Zhang, Vicente, Dideriksen, Arora, et~al.]{abo}
Jasmine Collins, Shubham Goel, Kenan Deng, Achleshwar Luthra, Leon Xu, Erhan Gundogdu, Xi Zhang, Tomas F~Yago Vicente, Thomas Dideriksen, Himanshu Arora, et~al.
\newblock Abo: Dataset and benchmarks for real-world 3d object understanding.
\newblock In \emph{CVPR}, 2022.

\bibitem[Dai et~al.(2017)Dai, Chang, Savva, Halber, Funkhouser, and Nie{\ss}ner]{scannet}
Angela Dai, Angel~X. Chang, Manolis Savva, Maciej Halber, Thomas Funkhouser, and Matthias Nie{\ss}ner.
\newblock Scannet: Richly-annotated 3d reconstructions of indoor scenes.
\newblock In \emph{CVPR}, 2017.

\bibitem[Deitke et~al.(2023)Deitke, Schwenk, Salvador, Weihs, Michel, VanderBilt, Schmidt, Ehsani, Kembhavi, and Farhadi]{objaverse}
Matt Deitke, Dustin Schwenk, Jordi Salvador, Luca Weihs, Oscar Michel, Eli VanderBilt, Ludwig Schmidt, Kiana Ehsani, Aniruddha Kembhavi, and Ali Farhadi.
\newblock Objaverse: A universe of annotated 3d objects.
\newblock In \emph{CVPR}, 2023.

\bibitem[Dhariwal and Nichol(2021)]{diffbeat}
Prafulla Dhariwal and Alex Nichol.
\newblock Diffusion models beat gans on image synthesis.
\newblock In \emph{NeurIPS}, 2021.

\bibitem[Duan et~al.(2025)Duan, Guo, and Zhu]{duan2025diffusiondepth}
Yiquan Duan, Xianda Guo, and Zheng Zhu.
\newblock Diffusiondepth: Diffusion denoising approach for monocular depth estimation.
\newblock In \emph{ECCV}, 2025.

\bibitem[Fang et~al.(2023)Fang, Wang, Fang, Gou, Liu, Yan, Liu, Xie, and Lu]{anygrasp}
Hao-Shu Fang, Chenxi Wang, Hongjie Fang, Minghao Gou, Jirong Liu, Hengxu Yan, Wenhai Liu, Yichen Xie, and Cewu Lu.
\newblock Anygrasp: Robust and efficient grasp perception in spatial and temporal domains.
\newblock \emph{T-RO}, 2023.

\bibitem[Fu et~al.(2021{\natexlab{a}})Fu, Cai, Gao, Zhang, Wang, Li, Zeng, Sun, Jia, Zhao, et~al.]{3dfront}
Huan Fu, Bowen Cai, Lin Gao, Ling-Xiao Zhang, Jiaming Wang, Cao Li, Qixun Zeng, Chengyue Sun, Rongfei Jia, Binqiang Zhao, et~al.
\newblock 3d-front: 3d furnished rooms with layouts and semantics.
\newblock In \emph{ICCV}, 2021{\natexlab{a}}.

\bibitem[Fu et~al.(2021{\natexlab{b}})Fu, Jia, Gao, Gong, Zhao, Maybank, and Tao]{future}
Huan Fu, Rongfei Jia, Lin Gao, Mingming Gong, Binqiang Zhao, Steve Maybank, and Dacheng Tao.
\newblock 3d-future: 3d furniture shape with texture.
\newblock \emph{IJCT}, 2021{\natexlab{b}}.

\bibitem[Goodfellow et~al.(2020)Goodfellow, Pouget-Abadie, Mirza, Xu, Warde-Farley, Ozair, Courville, and Bengio]{goodfellow2020generative}
Ian Goodfellow, Jean Pouget-Abadie, Mehdi Mirza, Bing Xu, David Warde-Farley, Sherjil Ozair, Aaron Courville, and Yoshua Bengio.
\newblock Generative adversarial networks.
\newblock \emph{Communications of the ACM}, 2020.

\bibitem[Gu et~al.(2020)Gu, Guo, Deligianni, and Yang]{gu2020coupled}
Xiao Gu, Yao Guo, Fani Deligianni, and Guang-Zhong Yang.
\newblock Coupled real-synthetic domain adaptation for real-world deep depth enhancement.
\newblock \emph{TIP}, 2020.

\bibitem[Ho et~al.(2020)Ho, Jain, and Abbeel]{ho2020denoising}
Jonathan Ho, Ajay Jain, and Pieter Abbeel.
\newblock Denoising diffusion probabilistic models.
\newblock In \emph{NeurIPS}, 2020.

\bibitem[Ho et~al.(2022)Ho, Saharia, Chan, Fleet, Norouzi, and Salimans]{ho2022cascaded}
Jonathan Ho, Chitwan Saharia, William Chan, David~J Fleet, Mohammad Norouzi, and Tim Salimans.
\newblock Cascaded diffusion models for high fidelity image generation.
\newblock \emph{Journal of Machine Learning Research}, 23\penalty0 (47):\penalty0 1--33, 2022.

\bibitem[Isola et~al.(2017)Isola, Zhu, Zhou, and Efros]{pix2pix}
Phillip Isola, Jun-Yan Zhu, Tinghui Zhou, and Alexei~A Efros.
\newblock Image-to-image translation with conditional adversarial networks.
\newblock In \emph{CVPR}, 2017.

\bibitem[Jensen et~al.(2014)Jensen, Dahl, Vogiatzis, Tola, and Aan{\ae}s]{dtu}
Rasmus Jensen, Anders Dahl, George Vogiatzis, Engil Tola, and Henrik Aan{\ae}s.
\newblock Large scale multi-view stereopsis evaluation.
\newblock In \emph{CVPR}, 2014.

\bibitem[Kang et~al.(2024)Kang, Chang, and Choi]{kang2024balanced}
Cheongwoong Kang, Wonjoon Chang, and Jaesik Choi.
\newblock Balanced domain randomization for safe reinforcement learning.
\newblock \emph{Applied Sciences}, 14\penalty0 (21):\penalty0 9710, 2024.

\bibitem[Kang et~al.(2023)Kang, Zhu, Zhang, Park, Shechtman, Paris, and Park]{kang2023gigagan}
Minguk Kang, Jun-Yan Zhu, Richard Zhang, Jaesik Park, Eli Shechtman, Sylvain Paris, and Taesung Park.
\newblock Scaling up gans for text-to-image synthesis.
\newblock In \emph{CVPR}, 2023.

\bibitem[Ke et~al.(2024)Ke, Obukhov, Huang, Metzger, Daudt, and Schindler]{ke2024repurposing}
Bingxin Ke, Anton Obukhov, Shengyu Huang, Nando Metzger, Rodrigo~Caye Daudt, and Konrad Schindler.
\newblock Repurposing diffusion-based image generators for monocular depth estimation.
\newblock In \emph{CVPR}, 2024.

\bibitem[Khan et~al.(2021)Khan, Kim, and Tompkin]{khan2021differentiable}
Numair Khan, Min~H Kim, and James Tompkin.
\newblock Differentiable diffusion for dense depth estimation from multi-view images.
\newblock In \emph{CVPR}, 2021.

\bibitem[Kingma and Welling(2014)]{kingma2014auto}
Diederik~P Kingma and Max Welling.
\newblock Auto-encoding variational bayes.
\newblock \emph{STAT}, 1050:\penalty0 1, 2014.

\bibitem[Kodali et~al.(2017)Kodali, Abernethy, Hays, and Kira]{gans}
Naveen Kodali, Jacob~D. Abernethy, James Hays, and Zsolt Kira.
\newblock On convergence and stability of gans.
\newblock \emph{arXiv preprint arXiv:1705.07215}, 2017.

\bibitem[Landau et~al.(2015)Landau, Choo, and Beling]{landau2015simulating}
Michael~J Landau, Benjamin~Y Choo, and Peter~A Beling.
\newblock Simulating kinect infrared and depth images.
\newblock \emph{IEEE transactions on cybernetics}, 2015.

\bibitem[Liu et~al.(2024)Liu, Ye, Nie, He, and Han]{liu2024lasa}
Haolin Liu, Chongjie Ye, Yinyu Nie, Yingfan He, and Xiaoguang Han.
\newblock Lasa: Instance reconstruction from real scans using a large-scale aligned shape annotation dataset.
\newblock In \emph{CVPR}, 2024.

\bibitem[Loshchilov and Hutter(2019)]{loshchilov2019decoupled}
Ilya Loshchilov and Frank Hutter.
\newblock Decoupled weight decay regularization.
\newblock \emph{arXiv preprint arXiv:1711.05101}, 2019.

\bibitem[Matsunaga et~al.(2022)Matsunaga, Ishii, Hayakawa, Suzuki, and Narihira]{matsunaga2022fine}
Naoki Matsunaga, Masato Ishii, Akio Hayakawa, Kenji Suzuki, and Takuya Narihira.
\newblock Fine-grained image editing by pixel-wise guidance using diffusion models.
\newblock \emph{arXiv preprint arXiv:2212.02024}, 2022.

\bibitem[Meister et~al.(2013)Meister, Nair, and Kondermann]{meister2013simulation}
Stephan Meister, Rahul Nair, and Daniel Kondermann.
\newblock Simulation of time-of-flight sensors using global illumination.
\newblock In \emph{VMV}, 2013.

\bibitem[Meng et~al.(2021)Meng, He, Song, Song, Wu, Zhu, and Ermon]{meng2021sdedit}
Chenlin Meng, Yutong He, Yang Song, Jiaming Song, Jiajun Wu, Jun-Yan Zhu, and Stefano Ermon.
\newblock Sdedit: Guided image synthesis and editing with stochastic differential equations.
\newblock In \emph{ICLR}, 2021.

\bibitem[Mo et~al.(2023)Mo, Xie, Chu, Yao, Hong, Nießner, and Li]{mo2023dit3d}
Shentong Mo, Enze Xie, Ruihang Chu, Lewei Yao, Lanqing Hong, Matthias Nießner, and Zhenguo Li.
\newblock Dit-3d: Exploring plain diffusion transformers for 3d shape generation.
\newblock \emph{arXiv preprint arXiv: 2307.01831}, 2023.

\bibitem[Nichol et~al.(2021)Nichol, Dhariwal, Ramesh, Shyam, Mishkin, McGrew, Sutskever, and Chen]{nichol2021glide}
Alex Nichol, Prafulla Dhariwal, Aditya Ramesh, Pranav Shyam, Pamela Mishkin, Bob McGrew, Ilya Sutskever, and Mark Chen.
\newblock Glide: Towards photorealistic image generation and editing with text-guided diffusion models.
\newblock \emph{arXiv preprint arXiv:2112.10741}, 2021.

\bibitem[Nichol and Dhariwal(2021)]{nichol2021improved}
Alexander~Quinn Nichol and Prafulla Dhariwal.
\newblock Improved denoising diffusion probabilistic models.
\newblock In \emph{ICML}, 2021.

\bibitem[Pashevich et~al.(2019)Pashevich, Strudel, Kalevatykh, Laptev, and Schmid]{pashevich2019learning}
Alexander Pashevich, Robin Strudel, Igor Kalevatykh, Ivan Laptev, and Cordelia Schmid.
\newblock Learning to augment synthetic images for sim2real policy transfer.
\newblock In \emph{IROS}, 2019.

\bibitem[Paszke et~al.(2019)Paszke, Gross, Massa, Lerer, Bradbury, Chanan, Killeen, Lin, Gimelshein, Antiga, Desmaison, K\"{o}pf, Yang, DeVito, Raison, Tejani, Chilamkurthy, Steiner, Fang, Bai, and Chintala]{pytorch}
Adam Paszke, Sam Gross, Francisco Massa, Adam Lerer, James Bradbury, Gregory Chanan, Trevor Killeen, Zeming Lin, Natalia Gimelshein, Luca Antiga, Alban Desmaison, Andreas K\"{o}pf, Edward Yang, Zach DeVito, Martin Raison, Alykhan Tejani, Sasank Chilamkurthy, Benoit Steiner, Lu Fang, Junjie Bai, and Soumith Chintala.
\newblock Pytorch: an imperative style, high-performance deep learning library.
\newblock In \emph{NeurIPS}, 2019.

\bibitem[Peng et~al.(2023)Peng, Hu, Ke, and Liu]{peng2023diffusion}
Duo Peng, Ping Hu, Qiuhong Ke, and Jun Liu.
\newblock Diffusion-based image translation with label guidance for domain adaptive semantic segmentation.
\newblock In \emph{ICCV}, 2023.

\bibitem[Planche et~al.(2017)Planche, Wu, Ma, Sun, Kluckner, Lehmann, Chen, Hutter, Zakharov, Kosch, et~al.]{planche2017depthsynth}
Benjamin Planche, Ziyan Wu, Kai Ma, Shanhui Sun, Stefan Kluckner, Oliver Lehmann, Terrence Chen, Andreas Hutter, Sergey Zakharov, Harald Kosch, et~al.
\newblock Depthsynth: Real-time realistic synthetic data generation from cad models for 2.5 d recognition.
\newblock In \emph{3DV}, 2017.

\bibitem[Qi et~al.(2017)Qi, Su, Mo, and Guibas]{qi2017pointnet}
Charles~R Qi, Hao Su, Kaichun Mo, and Leonidas~J Guibas.
\newblock Pointnet: Deep learning on point sets for 3d classification and segmentation.
\newblock In \emph{CVPR}, 2017.

\bibitem[Radford et~al.(2021)Radford, Kim, Hallacy, Ramesh, Goh, Agarwal, Sastry, Askell, Mishkin, Clark, Krueger, and Sutskever]{radford2021clip}
Alec Radford, Jong~Wook Kim, Chris Hallacy, Aditya Ramesh, Gabriel Goh, Sandhini Agarwal, Girish Sastry, Amanda Askell, Pamela Mishkin, Jack Clark, Gretchen Krueger, and Ilya Sutskever.
\newblock Learning transferable visual models from natural language supervision.
\newblock In \emph{ICML}, 2021.

\bibitem[Rombach et~al.(2022)Rombach, Blattmann, Lorenz, Esser, and Ommer]{stablediffusion}
Robin Rombach, Andreas Blattmann, Dominik Lorenz, Patrick Esser, and Bjorn Ommer.
\newblock High-resolution image synthesis with latent diffusion models.
\newblock In \emph{CVPR}, 2022.

\bibitem[Ruiz et~al.(2023)Ruiz, Li, Jampani, Pritch, Rubinstein, and Aberman]{ruiz2023dreambooth}
Nataniel Ruiz, Yuanzhen Li, Varun Jampani, Yael Pritch, Michael Rubinstein, and Kfir Aberman.
\newblock Dreambooth: Fine tuning text-to-image diffusion models for subject-driven generation.
\newblock In \emph{CVPR}, 2023.

\bibitem[Saharia et~al.(2022)Saharia, Ho, Chan, Salimans, Fleet, and Norouzi]{saharia2022image}
Chitwan Saharia, Jonathan Ho, William Chan, Tim Salimans, David~J Fleet, and Mohammad Norouzi.
\newblock Image super-resolution via iterative refinement.
\newblock \emph{TPAMI}, 2022.

\bibitem[Saxena et~al.(2023)Saxena, Kar, Norouzi, and Fleet]{saxena2023monocular}
Saurabh Saxena, Abhishek Kar, Mohammad Norouzi, and David~J Fleet.
\newblock Monocular depth estimation using diffusion models.
\newblock \emph{arXiv preprint arXiv:2302.14816}, 2023.

\bibitem[Saxena et~al.(2024)Saxena, Herrmann, Hur, Kar, Norouzi, Sun, and Fleet]{saxena2024surprising}
Saurabh Saxena, Charles Herrmann, Junhwa Hur, Abhishek Kar, Mohammad Norouzi, Deqing Sun, and David~J Fleet.
\newblock The surprising effectiveness of diffusion models for optical flow and monocular depth estimation.
\newblock \emph{NeurIPS}, 2024.

\bibitem[Shen et~al.(2022)Shen, Yang, Zheng, Liu, and Guibas]{shen2022dcl}
Yuefan Shen, Yanchao Yang, Youyi Zheng, C~Karen Liu, and Leonidas~J Guibas.
\newblock Dcl: Differential contrastive learning for geometry-aware depth synthesis.
\newblock \emph{IEEE Robotics and Automation Letters}, 2022.

\bibitem[Shi et~al.(2024)Shi, Cao, Xia, Zhu, Liao, and Yang]{shi2024dsr}
Yuan Shi, Huiyun Cao, Bin Xia, Rui Zhu, Qingmin Liao, and Wenming Yang.
\newblock Dsr-diff: Depth map super-resolution with diffusion model.
\newblock \emph{Pattern Recognition Letters}, 184:\penalty0 225--231, 2024.

\bibitem[Shrivastava et~al.(2017)Shrivastava, Pfister, Tuzel, Susskind, Wang, and Webb]{shrivastava2017learning}
Ashish Shrivastava, Tomas Pfister, Oncel Tuzel, Joshua Susskind, Wenda Wang, and Russell Webb.
\newblock Learning from simulated and unsupervised images through adversarial training.
\newblock In \emph{CVPR}, 2017.

\bibitem[Sohl-Dickstein et~al.(2015)Sohl-Dickstein, Weiss, Maheswaranathan, and Ganguli]{dpm2015}
Jascha Sohl-Dickstein, Eric~A. Weiss, Niru Maheswaranathan, and Surya Ganguli.
\newblock Deep unsupervised learning using nonequilibrium thermodynamics.
\newblock In \emph{ICML}, 2015.

\bibitem[Song et~al.(2020{\natexlab{a}})Song, Meng, and Ermon]{song2020denoising}
Jiaming Song, Chenlin Meng, and Stefano Ermon.
\newblock Denoising diffusion implicit models.
\newblock \emph{arXiv preprint arXiv:2010.02502}, 2020{\natexlab{a}}.

\bibitem[Song et~al.(2015)Song, Lichtenberg, and Xiao]{sunrgbd}
Shuran Song, Samuel~P. Lichtenberg, and Jianxiong Xiao.
\newblock Sun rgb-d: A rgb-d scene understanding benchmark suite.
\newblock In \emph{CVPR}, 2015.

\bibitem[Song and Ermon(2019)]{song2019generative}
Yang Song and Stefano Ermon.
\newblock Generative modeling by estimating gradients of the data distribution.
\newblock In \emph{NeurIPS}, 2019.

\bibitem[Song and Ermon(2020)]{song2020improved}
Yang Song and Stefano Ermon.
\newblock Improved techniques for training score-based generative models.
\newblock In \emph{NeurIPS}, 2020.

\bibitem[Song et~al.(2020{\natexlab{b}})Song, Sohl-Dickstein, Kingma, Kumar, Ermon, and Poole]{song2020score}
Yang Song, Jascha Sohl-Dickstein, Diederik~P Kingma, Abhishek Kumar, Stefano Ermon, and Ben Poole.
\newblock Score-based generative modeling through stochastic differential equations.
\newblock \emph{arXiv preprint arXiv:2011.13456}, 2020{\natexlab{b}}.

\bibitem[Sweeney et~al.(2019)Sweeney, Izatt, and Tedrake]{sweeney2019supervised}
Chris Sweeney, Greg Izatt, and Russ Tedrake.
\newblock A supervised approach to predicting noise in depth images.
\newblock In \emph{ICRA}, 2019.

\bibitem[Tobin et~al.(2017)Tobin, Fong, Ray, Schneider, Zaremba, and Abbeel]{random}
Josh Tobin, Rachel Fong, Alex Ray, Jonas Schneider, Wojciech Zaremba, and Pieter Abbeel.
\newblock Domain randomization for transferring deep neural networks from simulation to the real world.
\newblock In \emph{IROS}, 2017.

\bibitem[Tobin et~al.(2018)Tobin, Biewald, Duan, Andrychowicz, Handa, Kumar, McGrew, Ray, Schneider, Welinder, et~al.]{tobin2018domain}
Josh Tobin, Lukas Biewald, Rocky Duan, Marcin Andrychowicz, Ankur Handa, Vikash Kumar, Bob McGrew, Alex Ray, Jonas Schneider, Peter Welinder, et~al.
\newblock Domain randomization and generative models for robotic grasping.
\newblock In \emph{IROS}, 2018.

\bibitem[Tosi et~al.(2024)Tosi, Ramirez, and Poggi]{tosi2024diffusion}
Fabio Tosi, Pierluigi~Zama Ramirez, and Matteo Poggi.
\newblock Diffusion models for monocular depth estimation: Overcoming challenging conditions.
\newblock \emph{arXiv preprint arXiv:2407.16698}, 2024.

\bibitem[Tremblay et~al.(2018)Tremblay, Prakash, Acuna, Brophy, Jampani, Anil, To, Cameracci, Boochoon, and Birchfield]{tremblay2018training}
Jonathan Tremblay, Aayush Prakash, David Acuna, Mark Brophy, Varun Jampani, Cem Anil, Thang To, Eric Cameracci, Shaad Boochoon, and Stan Birchfield.
\newblock Training deep networks with synthetic data: Bridging the reality gap by domain randomization.
\newblock In \emph{CVPRW}, 2018.

\bibitem[Uy et~al.(2019)Uy, Pham, Hua, Nguyen, and Yeung]{scanobjectnn}
Mikaela~Angelina Uy, Quang-Hieu Pham, Binh-Son Hua, Duc~Thanh Nguyen, and Sai-Kit Yeung.
\newblock Revisiting point cloud classification: A new benchmark dataset and classification model on real-world data.
\newblock In \emph{ICCV}, 2019.

\bibitem[Vahdat et~al.(2021)Vahdat, Kreis, and Kautz]{vahdat2021score}
Arash Vahdat, Karsten Kreis, and Jan Kautz.
\newblock Score-based generative modeling in latent space.
\newblock In \emph{NeurIPS}, 2021.

\bibitem[Wang et~al.(2021)Wang, Liu, Yue, Lasenby, and Kusner]{OcCo}
Hanchen Wang, Qi Liu, Xiangyu Yue, Joan Lasenby, and Matthew~J. Kusner.
\newblock Unsupervised point cloud pre-training via occlusion completion.
\newblock In \emph{ICCV}, 2021.

\bibitem[Wang et~al.(2024)Wang, Lin, Nie, Liao, Shao, and Zhao]{wang2024digging}
Jiyuan Wang, Chunyu Lin, Lang Nie, Kang Liao, Shuwei Shao, and Yao Zhao.
\newblock Digging into contrastive learning for robust depth estimation with diffusion models.
\newblock In \emph{ACM MM}, 2024.

\bibitem[Wang et~al.(2020)Wang, Goldluecke, and Anklam]{wang2020l2r}
Leichen Wang, Bastian Goldluecke, and Carsten Anklam.
\newblock L2r gan: Lidar-to-radar translation.
\newblock In \emph{ACCV}, 2020.

\bibitem[Wang et~al.(2019)Wang, Sun, Liu, Sarma, Bronstein, and Solomon]{dgcnn}
Yue Wang, Yongbin Sun, Ziwei Liu, Sanjay~E. Sarma, Michael~M. Bronstein, and Justin~M. Solomon.
\newblock Dynamic graph cnn for learning on point clouds.
\newblock \emph{TOG}, 2019.

\bibitem[Wang et~al.(2025)Wang, Xu, Tan, Chai, Liu, Pandey, Fanello, Kadambi, and Zhang]{wang2025mvdd}
Zhen Wang, Qiangeng Xu, Feitong Tan, Menglei Chai, Shichen Liu, Rohit Pandey, Sean Fanello, Achuta Kadambi, and Yinda Zhang.
\newblock Mvdd: Multi-view depth diffusion models.
\newblock In \emph{ECCV}, 2025.

\bibitem[Wu et~al.(2015)Wu, Song, Khosla, Yu, Zhang, Tang, and Xiao]{modelnet}
Zhirong Wu, Shuran Song, Aditya Khosla, Fisher Yu, Linguang Zhang, Xiaoou Tang, and Jianxiong Xiao.
\newblock 3d shapenets: A deep representation for volumetric shapes.
\newblock In \emph{CVPR}, 2015.

\bibitem[Xu et~al.(2022)Xu, Chen, Liu, and Han]{toscene}
Mutian Xu, Pei Chen, Haolin Liu, and Xiaoguang Han.
\newblock To-scene: A large-scale dataset for understanding 3d tabletop scenes.
\newblock In \emph{ECCV}, 2022.

\bibitem[Yu et~al.(2023)Yu, Xu, Zhang, Liu, Ye, Wu, Yan, Liang, Chen, Cui, and Han]{mvimgnet}
Xianggang Yu, Mutian Xu, Yidan Zhang, Haolin Liu, Chongjie Ye, Yushuang Wu, Zizheng Yan, Tianyou Liang, Guanying Chen, Shuguang Cui, and Xiaoguang Han.
\newblock Mvimgnet: A large-scale dataset of multi-view images.
\newblock In \emph{CVPR}, 2023.

\bibitem[Yuan et~al.(2022)Yuan, Shi, Feng, Chang, Liu, Chen, Knoll, and Zhang]{yuan2022sim}
Chengjie Yuan, Yunlei Shi, Qian Feng, Chunyang Chang, Michael Liu, Zhaopeng Chen, Alois~Christian Knoll, and Jianwei Zhang.
\newblock Sim-to-real transfer of robotic assembly with visual inputs using cyclegan and force control.
\newblock In \emph{2022 IEEE International Conference on Robotics and Biomimetics (ROBIO)}, pages 1426--1432. IEEE, 2022.

\bibitem[Yue et~al.(2019)Yue, Zhang, Zhao, Sangiovanni-Vincentelli, Keutzer, and Gong]{yue2019domain}
Xiangyu Yue, Yang Zhang, Sicheng Zhao, Alberto Sangiovanni-Vincentelli, Kurt Keutzer, and Boqing Gong.
\newblock Domain randomization and pyramid consistency: Simulation-to-real generalization without accessing target domain data.
\newblock In \emph{Proceedings of the IEEE/CVF international conference on computer vision}, pages 2100--2110, 2019.

\bibitem[Zhang et~al.(2023{\natexlab{a}})Zhang, Rao, and Agrawala]{controlnet}
Lvmin Zhang, Anyi Rao, and Maneesh Agrawala.
\newblock Adding conditional control to text-to-image diffusion models.
\newblock In \emph{ICCV}, 2023{\natexlab{a}}.

\bibitem[Zhang et~al.(2023{\natexlab{b}})Zhang, Rao, and Agrawala]{zhang2023adding}
Lvmin Zhang, Anyi Rao, and Maneesh Agrawala.
\newblock Adding conditional control to text-to-image diffusion models.
\newblock In \emph{ICCV}, 2023{\natexlab{b}}.

\bibitem[Zhang et~al.(2023{\natexlab{c}})Zhang, Chen, Li, Xiang, Qin, Gu, Ling, Liu, Zeng, Han, Huang, Mu, Xu, and Su]{optical}
Xiaoshuai Zhang, Rui Chen, Ang Li, Fanbo Xiang, Yuzhe Qin, Jiayuan Gu, Zhan Ling, Minghua Liu, Peiyu Zeng, Songfang Han, Zhiao Huang, Tongzhou Mu, Jing Xu, and Hao Su.
\newblock Close the optical sensing domain gap by physics-grounded active stereo sensor simulation.
\newblock \emph{IEEE Transactions on Robotics}, 2023{\natexlab{c}}.

\bibitem[Zhang et~al.(2024)Zhang, Ke, Riemenschneider, Metzger, Obukhov, Gross, Schindler, and Schroers]{zhang2024betterdepth}
Xiang Zhang, Bingxin Ke, Hayko Riemenschneider, Nando Metzger, Anton Obukhov, Markus Gross, Konrad Schindler, and Christopher Schroers.
\newblock Betterdepth: Plug-and-play diffusion refiner for zero-shot monocular depth estimation.
\newblock \emph{arXiv preprint arXiv:2407.17952}, 2024.

\bibitem[Zhu et~al.(2017)Zhu, Park, Isola, and Efros]{cyclegan}
Jun-Yan Zhu, Taesung Park, Phillip Isola, and Alexei~A Efros.
\newblock Unpaired image-to-image translation using cycle-consistent adversarial networks.
\newblock In \emph{ICCV}, 2017.

\end{thebibliography}
}

%%%%%%%%% Supplementary Material
\clearpage
%\appendix
\setcounter{figure}{0}
\setcounter{table}{0}

\begin{appendices}
% \section{\large{\textit{Supplement} of MVImgNet}}
\startcontents[supple]

{
    \hypersetup{linkcolor=black}
    \printcontents[supple]{}{1}{}
}
\renewcommand{\thefootnote}{\fnsymbol{footnote}}
%\usepackage{url}
% % \counterwithin{figure}{section}
% \counterwithin{table}{section}
\renewcommand{\thesection}{\Alph{section}}%
\renewcommand\thetable{\Roman{table}}
\renewcommand\thefigure{\Roman{figure}}

\section{More Qualitative Results }\label{sec:more_results}

\begin{figure*}[htbp]
    \centering
    \captionsetup{type=figure}
    \includegraphics[width=1.0\textwidth]{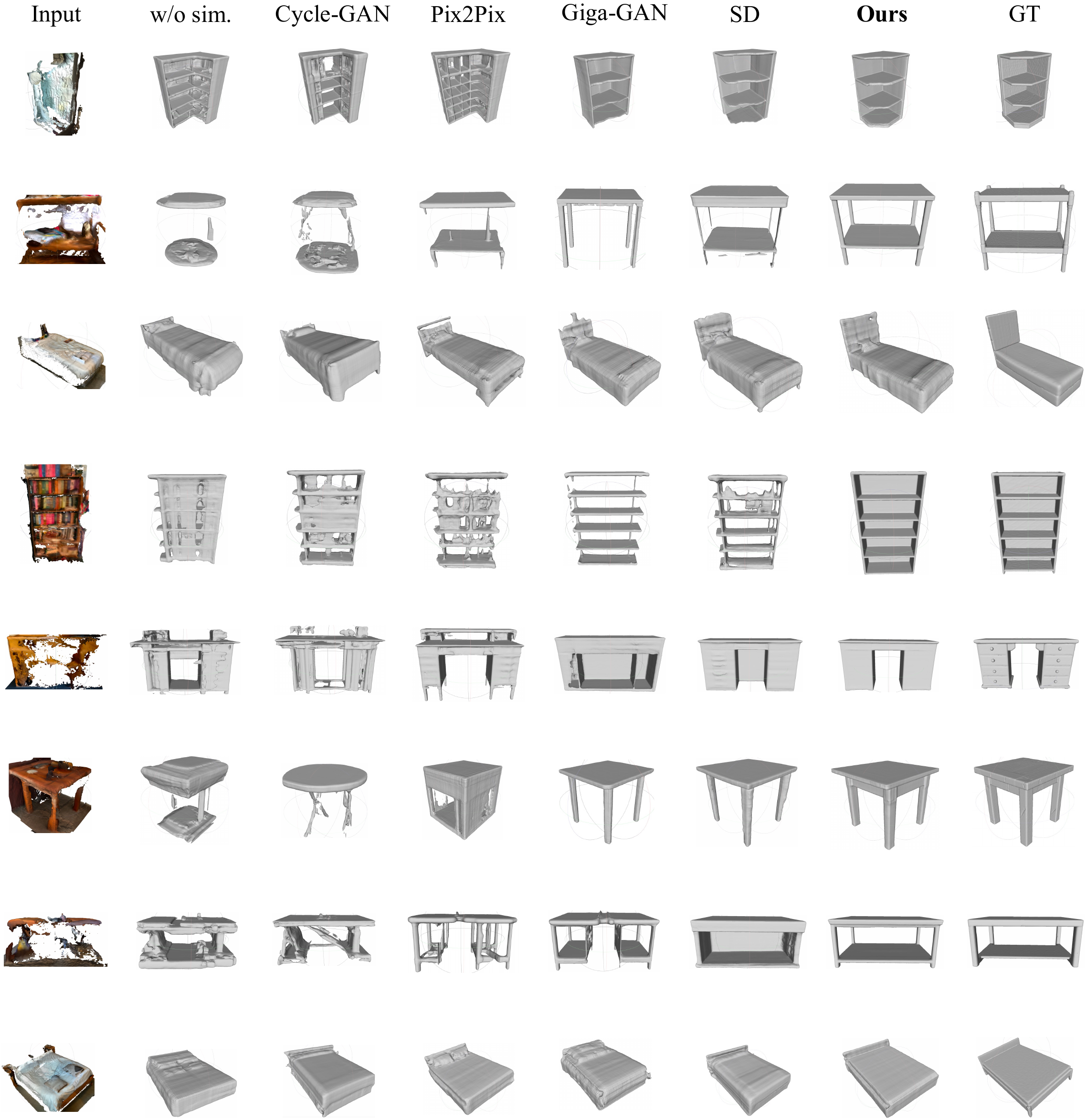}
    \captionof{figure}{Qualitative comparison on 3D instance reconstruction, using LASA \cite{liu2024lasa} training data mixed with simulated data generated by different methods. “Rand.” means random perturbation. “w/o sim.” denotes using LASA training data without simulated data.} \label{fig:qulitative_lasa_recon_supple}
    % \vspace{-0.3cm}
\end{figure*}%

\subsection{3D Instance Reconstruction}
\cref{fig:qulitative_lasa_recon_supple} presents more qualitative comparisons of 3D instance reconstruction on LASA validation set.
% The trained reconstruction model is tested on LASA validation set.

Consequently, our method consistently achieves remarkable 3D instance reconstruction results across diverse objects, from coarse to fine geometry.
% Notably, our output significantly improves reconstruction performance compared to using only the LASA training data (“w/o sim”). Furthermore, our Stage I output can achieve similar or slightly better performance than solely using LASA data. In contrast, other data simulation techniques noticeably reduce performance, due to their lower simulation quality, which introduces considerable ambiguity and complexity during the training of the reconstruction network. 
Moreover, given the reconstruction model DisCo is a diffusion-based model, we observe that our simulated data can enhance DisCo's proficiency in generating novel yet plausible geometries, such as the creation of a blanket on a bed.

% We provide a 3D animated visualization of the qualitative comparison in a visually appealing video format, where we rotate the reconstructed mesh. You may refer to the supplementary file folder and access the video file named \textcolor{red}{\textit{lasa\_reconstruction.mp4}} to view our qualitative results.

\subsection{Depth/3D Simulation}
\begin{figure}[t]
    \centering
    \captionsetup{type=figure}
    \includegraphics[width=1.0\linewidth]{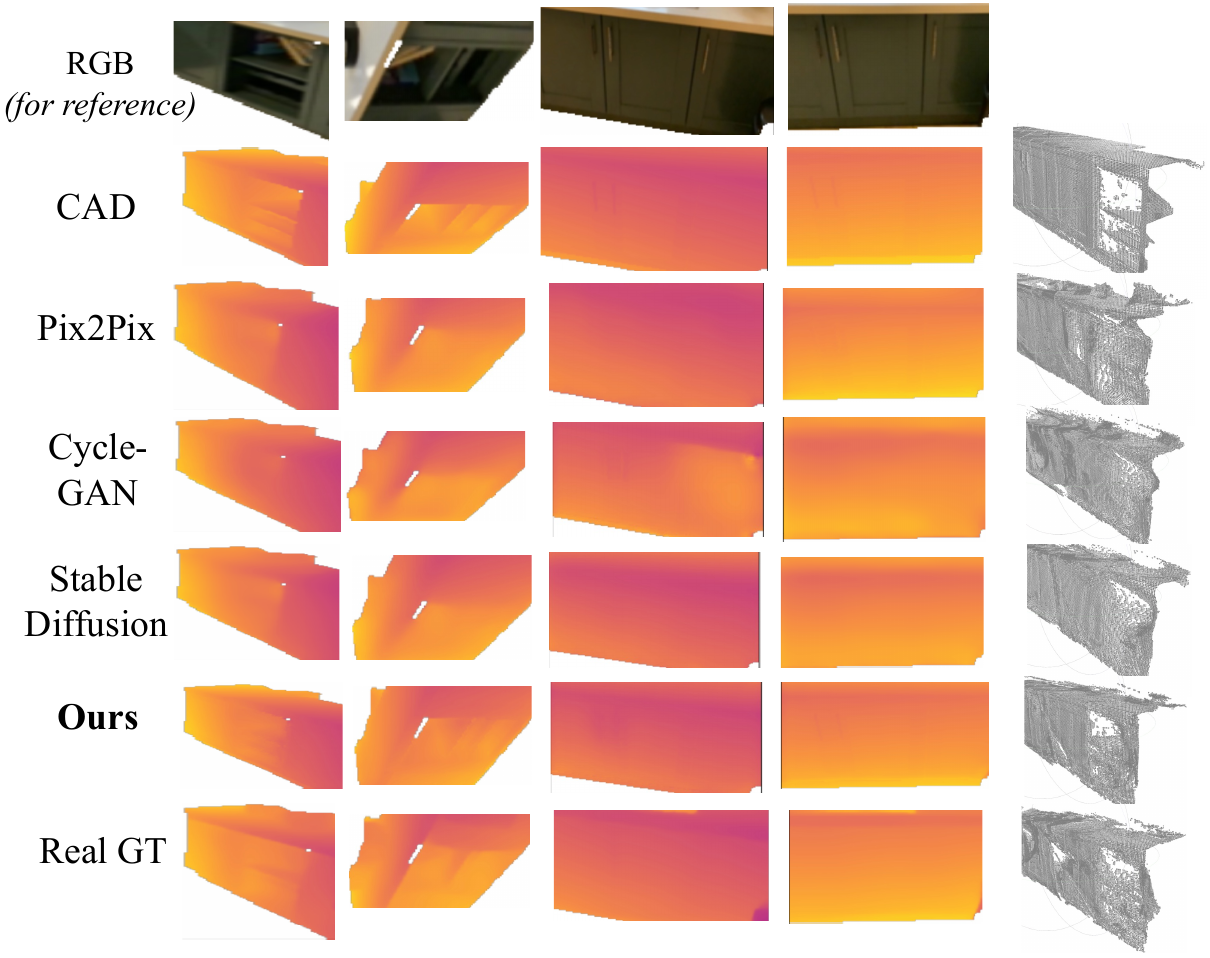}
    \captionof{figure}{Qualitative comparison of the depth/3D simulation results on LASA \cite{liu2024lasa} validation set.}
    \label{fig:qulitative_lasa_sim_supple}
    % \vspace{-0.4cm}
\end{figure}

In \cref{fig:qulitative_lasa_sim_supple}, we present additional qualitative results of depth/3D simulation on LASA \cite{liu2024lasa} validation set.
% he 3D simulated data generated by our method, in comparison to other approaches such as Cycle-GAN \cite{cyclegan} and Pix2Pix \cite{pix2pix}, closely resembles real captures, also exhibiting subtle distinctions from the real-captured ground truth (“Real GT”), particularly in fine-grained geometries, suggesting the rationality and diversity of our simulated data. Conversely, both Cycle-GAN \cite{cyclegan} and Pix2Pix \cite{pix2pix} introduce noticeably ambiguous distortions that \textit{deviate significantly} from real-world captures.
% The 3D animated visualization of the qualitative comparison is presented in \textcolor{red}{\textit{lasa\_sim2real.mp4}}.

\begin{figure}[t]
    \centering
    \captionsetup{type=figure}
    \includegraphics[width=1.0\linewidth]{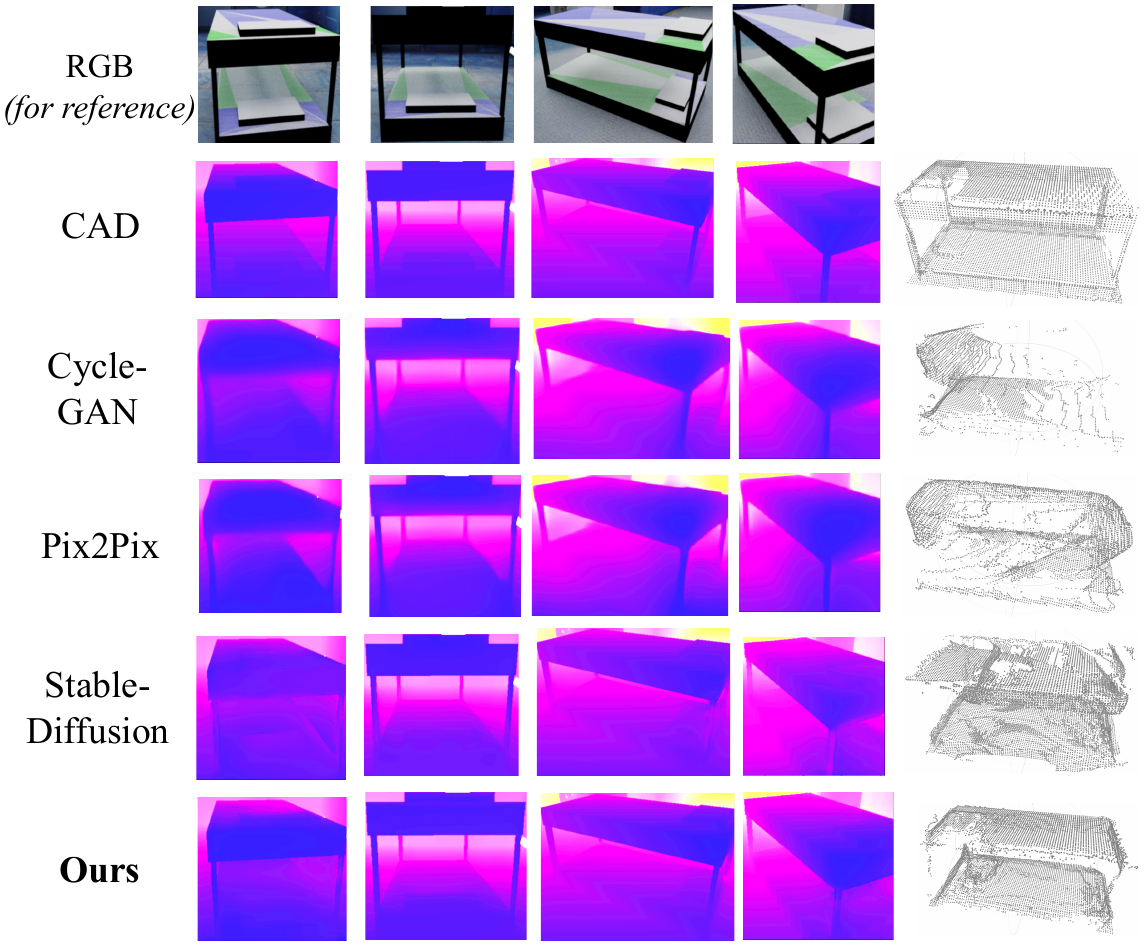}
    \captionof{figure}{Qualitative comparison of the depth/3D simulation results on ShapeNet \cite{shapenet}.}
    \label{fig:qulitative_syn_sim_supple}
    % \vspace{-0.4cm}
\end{figure}

Furthermore, \cref{fig:qulitative_syn_sim_supple} shows more depth/3D simulation results on ShapeNet \cite{shapenet}. 
% Similarly, Cycle-GAN and Pix2Pix \textit{severely distort} the original object geometries, whereas our method introduces more reasonable perturbations. As a result, the simulated data produced by Cycle-GAN and Pix2Pix introduce substantial \textit{ambiguity} during the training of reconstruction networks, leading to a decline in model performance, as depicted in Figure \ref{fig:qulitative_lasa_recon_supple}.
% The 3D animated visualization of the qualitative comparison is presented in \textcolor{red}{\textit{shapenet\_sim2real.mp4}}.

\section{User Study of Sim2Real}\label{sec:user_study}
To supplement the evaluation of our 3D simulation quality in the main paper, we conduct a user study. 
We invited 17 subjects through an online questionnaire. 
All participants are \textit{experienced} computer vision researchers, including senior professors or students. 
Half of them are experts in 3D vision, and none of them had seen our results before.
To begin with, we first randomly select a set of 20 real-captured ground truths (GT) from LASA \cite{liu2024lasa} validation set in 3D format, with the synthetic data as a reference, and instruct the participants to get familiar with the data pattern of GT.

During the evaluation, we randomly chose 10 samples each from the simulated data produced by Cycle-GAN \cite{cyclegan}, Pix2Pix \cite{pix2pix}, Giga-GAN \cite{kang2023gigagan}, Stable-Diffusion \cite{stablediffusion} our method, and the real-captured GT.
Subsequently, these 60 (10x6) samples were mixed together and presented to the participants.
Each participant was tasked to \textit{pick} the samples they perceived to be real captures based on their impressions and understanding of the initial samples.
% show the 3D simulated data generated by our method and the corresponding GT, arranged side by side in random order, also with the synthetic data as a reference. 
% Then we ask the subject to discriminate whether the xxxx.
For each method, we calculate the ratio of the quantity picked as real captures among all the presented samples produced by these methods.
As shown in \cref{tab:user_study}, our method surpasses other methods and approximates the result of real GT.

\begin{table}[t]
  \setlength{\tabcolsep}{3.0pt}
  \centering
  \resizebox{1.0\linewidth}{!}{
  \begin{tabular}{c|cccccc}
    \toprule
    Method & Cycle-GAN~\cite{cyclegan} & Pix2Pix~\cite{pix2pix} & Giga-GAN \cite{kang2023gigagan} & SD \cite{stablediffusion} & Ours & Real GT\\
    \midrule
    Picked ratio (\%)  & 22.5 & 33.1 & 55.4 & 62.7 & 88.6 & 94.8\\
    \bottomrule
  \end{tabular}}
  % \vspace{-0.2cm}
  \caption{The quantitative results of \textbf{user study}.}
  \label{tab:user_study}
  % \vspace{-0.4cm}
\end{table}

% \section{Depth Enhancement}\label{sec:depth_enhancement}

\section{More Ablation Studies}\label{sec:more_ablation}
We follow the main paper to conduct ablation studies on the 3D instance reconstruction task, specifically on the table category, using the zero-shot evaluation scheme.

\paragraph{3D binary classifier.}
We evaluate two key factors in the implementation of our 3D binary classifier, the size of split patches in the depth map and the network channel of the point classifier. 

\cref{tab:bin_ablation} reports the results using different patch sizes, where the classification accuracy of the binary classifier is also included. The best results are achieved with a 128x128 patch size, while using 64x64 or 256x256 sizes marginally reduces performance. Employing a 32x32 patch size results in very small projected point patches, making it challenging for the binary classifier to distinguish effectively, ultimately resulting in the poorest performance.

\begin{table}[t]
  \setlength{\tabcolsep}{3.0pt}
  \centering
  \resizebox{1.0\linewidth}{!}{
  \begin{tabular}{c|cccc}
    \toprule
    Patch size & 32x32 & 64x64 & \textbf{128x128} & 256x256\\
    \midrule
    Acc (\%) & 42.7 & 65.4 & 67.8 & 70.2 \\
    \midrule
    Metrics & 27.1/16.9/22.6 & 28.3/16.4/24.3 & \textbf{28.5}/
    \textbf{16.1}/\textbf{24.7} & 28.2/16.5/24.4\\
    \bottomrule
  \end{tabular}}
  % \vspace{-0.2cm}
  \caption{Quantitative ablations of the patch size to split the depth map. Evaluation metrics are \textbf{mIoU / Chamfer L2 / F-score} of 3D instance reconstruction respectively.}
  \label{tab:bin_ablation}
  % \vspace{-0.3cm}
\end{table}

\cref{tab:wid_ablation} presents the results of using different channels (\ie, network width of each layer) in the 3D binary classifier (\ie, PointNet \cite{qi2017pointnet}). Similarly, the binary classification accuracy is also reported. Consequently, setting the width to 64 yields the best result, while using 32 or 128 slightly degrades the performance. However, using either 16 or 256 may build a very weak or strong classifier, ultimately leading to inferior results.

\begin{table}[t]
  \setlength{\tabcolsep}{3.0pt}
  \centering
  \resizebox{1.0\linewidth}{!}{
  \begin{tabular}{c|ccccc}
    \toprule
    Width & 16 & 32 & \textbf{64} & 128 & 256\\
    \midrule
    Acc (\%) & 41.5 & 59.8 & 67.8 & 78.5 & 88.6\\
    \midrule
    Metrics & 26.3/17.2/22.8 & 28.3/16.4/24.2 & \textbf{28.5}/
    \textbf{16.1}/\textbf{24.7} & 28.1/16.7/23.9 & 26.4/18.3/22.5\\
    \bottomrule
  \end{tabular}}
  % \vspace{-0.2cm}
  \caption{Quantitative ablations of the network width (“Width”) in the 3D binary classifier. Evaluation metrics are \textbf{mIoU / Chamfer L2 / F-score} of 3D instance reconstruction respectively.}
  \label{tab:wid_ablation}
  % \vspace{-0.3cm}
\end{table}

\paragraph{Re-weighting diffusion loss.}
In Stage II of our framework, we re-weight the diffusion loss to prioritize the optimization of unsatisfactory areas. $\omega$ and $\lambda$ are weight coefficients of similar regions and distinct (\ie, unsatisfactory) areas, respectively, in which $\lambda > \omega$. 
\cref{tab:loss_weight_ablation} presents the ablations of these two coefficients. Setting $\omega=0.5$ and $\lambda=1.5$ results in the optimal outcome, while other reasonable configurations also consistently deliver good results.
Another alternative way is to use the output confidence value from the binary classifier as the soft weight. However, we find that this operation may degrade the performance (\cref{tab:loss_weight_ablation}: Conf.), which is probably due to its \textit{less control} over the loss weight compared to directly setting weight.

\begin{table}[t]
  \setlength{\tabcolsep}{3.0pt}
  \centering
  \resizebox{1.0\linewidth}{!}{
  \begin{tabular}{c|cccc|c}
    \toprule
    $\omega/\lambda$ & 0.3/1.7 & 0.4/1.6 & \textbf{0.5/1.5} & 0.6/1.4 & Conf.\\
    \midrule
    Metrics & 28.1/\textbf{16.1}/24.5 
    & 28.4/16.5/\textbf{24.7} 
    & 28.5/\textbf{16.1}/\textbf{24.7} 
    & \textbf{28.8}/15.9/24.3 
    & 27.8/15.2/23.9\\
    \bottomrule
  \end{tabular}}
  % \vspace{-0.2cm}
  \caption{Quantitative ablations of the weight coefficients to re-weight denoising loss. Evaluation metrics are \textbf{mIoU / Chamfer L2 / F-score} of 3D instance reconstruction respectively.}
  \label{tab:loss_weight_ablation}
  % \vspace{-0.3cm}
\end{table}

\paragraph{Why simulating 3D via 2D?}
In fact, our pipeline exactly \textit{mimics the real 3D data acquisition} where individual 2D depth maps (with noise) from different views are captured and fused into 3D.
As for depth fusion, since our method simulates and adds \textit{noise} to \textit{view-consistent} rendered synthetic depth inputs, the view-variation of the resulting simulated depth maps \textit{remains small}, making the simulated depth can always be effectively fused into 3D data in our experiment.

We also finetuned a recent 3D diffusion, DiT-3D \cite{mo2023dit3d} $^{\ref{footnote:dit3d_repo}}$ to simulate 3D data for 3D instance reconstruction. It gets 26.9/19.1/22.3 mIoU/CD/F-Score on the table category under the zero-shot evaluation scheme, which is much worse than our method (28.5/16.1/24.7). This verifies the necessity to leverage 2D priors to mitigate the issue of 3D data scarcity.

\footnotetext[1]{\label{footnote:dit3d_repo}\url{https://github.com/DiT-3D/DiT-3D}}

\footnotetext[2]{\label{footnote:gan_repo}\url{https://github.com/junyanz/pytorch-CycleGAN-and-pix2pix}}

\footnotetext[3]{\label{footnote:gigagan_repo}\url{https://github.com/lucidrains/gigagan-pytorch}}

\footnotetext[4]{\label{footnote:sd_repo}\url{https://huggingface.co/stabilityai/stable-diffusion-2-1}}

\section{Network and Implementation Details}\label{sec:implementation}

\paragraph{Details of different simulation methods.}

For both Cycle-GAN \cite{cyclegan} and Pix2Pix \cite{pix2pix}, we use their official PyTorch \cite{pytorch} code $^{\ref{footnote:gan_repo}}$, similarly for Giga-GAN $^{\ref{footnote:gigagan_repo}}$ and Stable-Diffusion $^{\ref{footnote:sd_repo}}$.
In our approach, in both Stage I and II, we fine-tune the pre-trained Stable Diffusion V2.1 $^{\ref{footnote:sd_repo}}$ \cite{stablediffusion} using the AdamW optimizer \cite{loshchilov2019decoupled} with a fixed learning rate of 3e-5. To align with the VAE's expected input range, we normalize all input maps to the range [-1, 1]. During training, we employ random crops with various aspect ratios and pad the images to a resolution of 512x512 using black padding. The batch size is set to 48 for approximately 20,000 steps. The entire training process spans about two days utilizing four A100 GPUs. Our network architecture is based on the building blocks of ControlNet \cite{controlnet}.

\paragraph{Details of 3D Binary Classifier.}
We use the original PointNet \cite{qi2017pointnet} as our 3D binary classifier, which is a network with several multi-layer-perceptrons (MLPs) with a max-pooling operation to extract the final global feature vector.  During training, we also extract the feature of synthetic point patches using one single-layer MLP and a max-pooling layer. To provide guidance, the extracted synthetic features are concatenated with the global feature of real-captured or generated point patches, forming the final feature for optimization and prediction. The training strategy also follows PointNet's original setting.

\section{More Discussions}\label{sec:more_discussion}

\subsection{Potentials}

\paragraph{Using our model for: denoise VS add-noise.}
Similar to the setting of conventional 2D tasks, we further apply our model to take noisy depth as inputs and \textit{\textbf{de}noise} them into clean depth, and find that using one-stage diffusion is \textit{sufficient} (PSNR: One-stage 42.9 VS Two-stage 43.2).
To clarify, as highlighted in the main paper, different from conventional 2D tasks which often remove noise from the input into a clean output, our special task poses a \textit{\textbf{unique challenge}} about adding noise to generate noisy output. To this end, our method is \textit{customized} to tackle this challenge, and applying it to traditional 2D tasks may \textit{\textbf{not} fully verify} its unique value.
Conversely, our work potentially opens a new task of simulating real-world image/RGB noise/degradation.

\paragraph{Still benefit denoise.}
From another perspective of data, our clean-to-noisy method exactly \textit{serves to \textbf{improve}} the inverse solution of noisy-to-clean: To train a noisy-to-clean model, it still requires large-scale clean-noisy \textit{paired data}. To gain/scale up such paired data, our method actually serves as a reasonable solution, which generates real/noisy data (\textit{\textbf{hard} to collect}) from synthetic input (\textit{\textbf{easy} to gain}).
In fact, this logic has been verified by our experiment of 3D instance reconstruction, where our clean-to-noisy method generates clean-noisy paired data that are used to train a better noisy-to-clean (\ie, recovering clean surface from noisy observation) model.
Our method fills/complements the \textit{closed loop} of clean-noisy-clean.

\subsection{Limitations}
As our model is finetuned on a fixed dataset (\ie, LASA \cite{liu2024lasa}), transferring our model to other sensor data or new domains of objects/scenes that have a distinct gap with LASA may require re-training the model currently, and we consider this as an open question for future exploration. Additionally, we expect new paired datasets to explore our data-driven simulation in the future.

Moreover, the current two-stage pipeline is complex, and a \textit{one-stage} pipeline could be achieved by training the \textit{generator} and our 3D \textit{classifier} \textit{simultaneously} in an \textit{adversarial} manner. We leave this for future work.

Last, our model relies on \textit{paired} data, yet we believe strong priors inherited from large pretrained models may reduce such reliance. This remains an open problem for future work.

\stopcontents[supple]

\end{appendices}

\end{document}